\documentclass[twoside]{article}
\usepackage{graphicx} 
\usepackage[style=ieee, maxbibnames=99, citestyle=numeric-comp, url=false]{biblatex}
\usepackage{wrapfig} 
\usepackage{amssymb} 
\usepackage{xcolor} 
\newcommand{\myred}[1]{\textcolor{red}{#1}}

\newcommand{\myblue}[1]{\textcolor{blue}{#1}}

\usepackage[sc]{mathpazo} 
\usepackage[T1]{fontenc} 
\usepackage{microtype} 
\usepackage[indent = 20pt]{parskip}

\usepackage[hmarginratio=1:1,top=32mm,columnsep=20pt]{geometry} 
\usepackage{multicol} 
\usepackage[hang, small,labelfont=bf,up,textfont=it,up]{caption} 
\usepackage{booktabs} 
\usepackage{float} 
	\usepackage{amsmath}
	\usepackage[hidelinks]{hyperref} 
	
	\usepackage{lettrine} 
	\usepackage{paralist} 
	
	\usepackage{abstract} 
	
	\usepackage{titlesec} 
	\titleformat{\section}[block]{\large\scshape\centering}{\thesection.}{1em}{} 
	\titleformat{\subsection}[block]{\large}{\thesubsection.}{1em}{} 
	
	\usepackage{fancyhdr} 
	\title{\vspace{-15mm}\fontsize{20pt}{10pt}\selectfont\textbf{A Data-Based Architecture for Flight Test without Test Points}} 
	
	\author{
		\textsc{D. Isaiah Harp}\thanks{Major, USAF; Experimental Test Pilot}\\[2mm] 
		\normalsize Principal Investigator, The Pointless Project \\ 
		\normalsize US Air Force Academy \\
		\normalsize \href{mailto:daniel.harp@afacademy.af.edu}{daniel.harp@afacademy.af.edu} 
		\and 
		\textsc{Joshua Ott}\thanks{1st Lieutenant, USAF; PhD Candidate}\\[2mm] 
		\normalsize Co-Investigator \\ 
		\normalsize Stanford University \\
		\normalsize \href{mailto:joshuaott@stanford.edu}{joshuaott@stanford.edu} 
		\and 
		\textsc{John Alora}\thanks{Major, USAF; PhD Candidate}\\[2mm] 
		\normalsize Stanford University \\
		\normalsize \href{mailto:jjalora@stanford.edu}{jjalora@stanford.edu} 
		\vspace{-5mm}
		\and 
		\textsc{Dylan M. Asmar}\thanks{Major, USAF; PhD Candidate}\\[2mm] 
		\normalsize Stanford University \\
		\normalsize \href{mailto:asmar@stanford.edu}{asmar@stanford.edu} 
		\vspace{-5mm}
	}

	\date{}
	
\begin{document}
		
		\maketitle 
		\thispagestyle{fancy} 
		
		
			
			\section{Motivation} \label{Motivation}
			
			In 2024, the Air Force executes envelope expansion flight test using the same experimental methods that it used in 1964. This stasis reflects a staggering rejection of modern computational methods. Sixty years ago, models were primarily analytical and generated predictions by hand or with rudimentary calculation machines. Today, high fidelity models (HFMs) that directly solve governing differential equations are routinely used to generate predictions of physical systems with tens of millions of degrees of freedom \cite{willcox2022data}. The exponential growth in computational power over the last six decades has clearly transformed engineering design and analysis. Yet the methods of test execution have not co-evolved.
			
			For example, envelope expansion for a major Air Force program lasted approximately six years. In the single discipline of structures, the test plan required 2199 test points to envelope the structural response of the air vehicle. During the test program, 13500 maneuvers were actually flown to complete these test points \cite{wendy2023rpg}. This implies that on average every point had to be repeated about six times to achieve acceptable tolerances and data quality. Only 16\% of the test execution yielded usable results. Assuming each maneuver required approximately the same amount of time, this analysis suggests that envelope expansion for this aircraft could have been completed in one year if every maneuver had generated usable data. 
			
			A full five years of this acquisition category one program were dedicated to the pursuit of unusable data.
			
			Past studies of the causes for delays in major acquisitions programs largely overlook this glaring inefficiency, because they only assess deviations from planned performance. The developmental test community builds these lengthy envelope expansion programs directly into their plans. Thus, the inefficiencies endemic to developmental test methods never appear as program delays. But streamlining these onerous methods is in fact fundamental to accelerating the acquisition of future air vehicles. Modern data-based engineering provides a tapestry of methods useful for so streamlining envelope expansion. Adopting them will get aircraft to the operators faster. 
			
			In this paper, we present an architecture for incorporating these data-based methods, including machine learning, into envelope expansion. This architecture succeeds by inverting the relationship between a model and the flight test data that validates it. With the data given theoretical primacy in this relationship, test pilots no longer must hit test points precisely. Envelope expansion becomes “point-less.” More importantly, the outcome of test becomes a compact representation of the underlying physics of the aircraft.

			To present this architecture, we shall first clarify and exemplify the contemporary flight test relationship between the model and data. We do so by reviewing two independent models (one analytical, one computational) that make predictions about flight test data. Second, we will propose a novel architecture that makes the data more fundamental than the model. Such an architecture has many advantages, which we will enumerate. Third, we will present an actual flight test campaign in which we demonstrated our architecture using a data-based method known as Gaussian Process Regression and estimated the short period characteristics for the T-38C. We compare our results with historical estimates of these parameters. We conclude with a future vision for our “Pointless Project.”

			\section{Background}
			
			\subsection{Model-Test-Validate} \label{MTV}
			
\begin{wrapfigure}{R}{0.6\textwidth}
					\centering
					\includegraphics[width=0.6\textwidth]{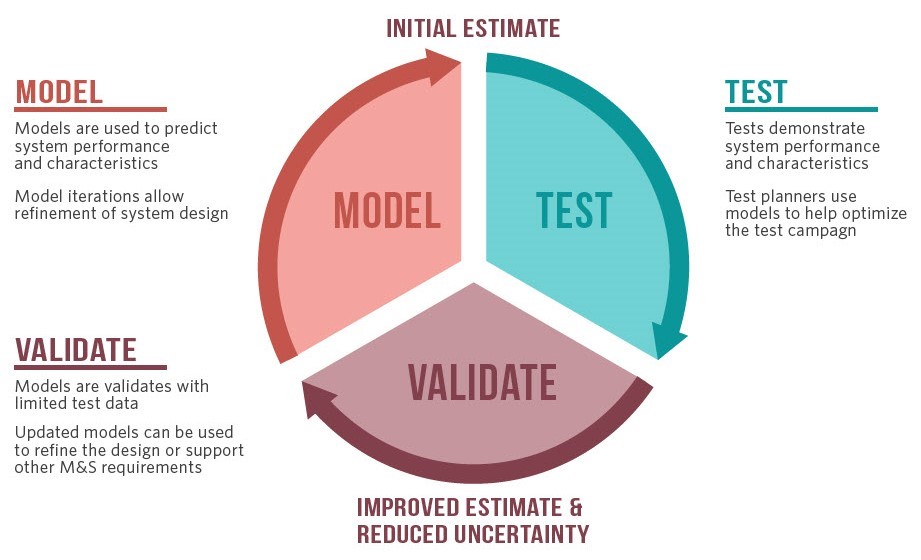}
					\caption{The Model-Test-Validate Cycle} \label{MTV-fig}
			\end{wrapfigure}

			In Defense Acquisition Magazine, Dr Eileen Bjorkman presented the model-test-validate cycle used in the Air Force Test Center as a graphic, which we reproduce in figure \ref{MTV-fig} \cite{bjorkman2023data}. This cycle is more commonly called the predict-test-validate cycle, but the use of “model” here is instructive: it emphasizes that predictions are always made by some model of the aircraft under test. In contemporary flight test, engineers first build a model of the aircraft. Then, they generate predictions by running that model. The activity of test generates data, which are then compared with the predictions generated by the model. This comparison is the fundamental outcome of flight test.

			The customary hierarchy – model primary, data substantiative – has two important consequences. 
			
			In the first place, it obligates test pilots to a high degree of precision. Models must be run with some particular set of assumptions: the parameters of the model. As those parameters change, the model outputs also change, sometimes in highly non-linear or poorly understood ways. Thus, to reproduce the predictions of the model the test pilot must first faithfully reproduce in the air the parameters of the model run. Especially in high-performance aircraft, this often requires the pilot to simultaneously achieve a multi-dimensional corner of Mach, dynamic pressure, altitude, and G. If the pilot deviates from the parameters used to produce the model, the data produced has no value. This is a weakness of the field of flight test, as data rejected for parametric unsuitability are just as informationally dense as data accepted. Yet the state of the art is to discard these data and repeat the point.
			
			The second consequence is more fundamental. Because the outcome of flight test amounts to reproductions of model predictions, each test point provides a local check of the model. Ergo, \emph{the fundamental outcome of developmental flight test is a series of spot-checks of a set of models.}\footnote{By “spot-check” we do not intend the sense of a “cursory inspection” but of a highly local inspection. We check the model as thoroughly as we can, but only at one particular spot.} \footnote{The reader might imagine this rather bold assertion to be falsified by flight test activities aimed at the determination of a quantity not modeled well prior to test. For example, does takeoff testing not determine the aircraft takeoff distance, suggesting that the output of the test was a value (takeoff distance) of direct use to the pilot? No, that’s not quite right. An individual takeoff distance measurement has no predictive capacity. We must perform many such measurements. As we do so, we implicitly build a statistical model of takeoff performance. This is a type of model, even if it’s just a Normal distribution; each data point samples (spot-checks) the distribution. More compellingly, these data will eventually be standardized and compiled into a chart that the pilot can use to predict takeoff distance on some given day. This chart is a model of takeoff performance, and every flight test measurement made has provided a validating spot-check of this model. That is, in fact, all flight test data has ever done.}
			
			The fact that flight test data only spot-checks models suggests a disconnect between desired and acquired knowledge. The tester and user both desire to understand the underlying physics of the aircraft in order, ultimately, to control it. But their empirical methods acquire only highly local inspections of some quantities that the underlying physics yields up to inspection. Like blind men who want to discern the shape of the elephant, we check first the tail, then the trunk, then the feet. Such a piece-wise inquiry is no doubt useful, but it is slow. The elephant will eventually come into view. But to advance, flight test must adopt methods that continuously view the whole. We claim that \emph{the fundamental output of developmental flight test ought to be a direct representation of the physics of the system under test.} 
   			
			We shall propose a flight test architecture that yields just such an output. By connecting data to physics, it responds to Bjorkman's declamation that test "is no longer just a data source for validating individual models \ldots it’s the knowledge source."\cite{bjorkman2023data}
   
            But first we clarify the current relationship between data and model with a motivating example, approached in two different ways: analytically, then computationally. 

			\subsection{A Motivating Example}
			
			Our architecture will generalize fully to the integrated disciplines of envelope expansion. However, to concretize its presentation we will rely on a classic problem in the single discipline of flying qualities: the prediction of the frequency and damping of the short period mode.
			
			Toward that end, consider a rigid aircraft with a single plane of symmetry. The longitudinal dynamics of this aircraft in the body frame can be expressed with the following equations:
			
			\begin{align}
				m(\dot{U}+QW-RV) &=\sum F_{x} \equiv \bar{X} \label{Xbar} \\
				m(\dot{W}+PV-QU) &=\sum F_{z} \equiv \bar{Z} \label{Zbar} \\
				I_{y}\dot{Q}+(I_{x}-I_{z})PR+I_{xz}(P^{2}-R^{2}) &=\sum M_{y} \equiv \bar{M} \label{Mbar}
			\end{align}
			
			Here, we have largely used the formulation and notation of Yechout in \cite{yechout2003introduction}; we will continue to do so throughout this paper except where we clearly define otherwise. These fully non-linear equations of motion can yield complex, coupled motions not expressible with closed-form solutions. However, these motions are dominated by a few dynamical modes. In particular, the longitudinal response of the aircraft angle-of-attack in response to elevator input – the “short period” – strongly governs how the aircraft behaves in pitch \cite{yechout2003introduction}. Historically, various modeling and flight test techniques have been employed to yield estimates of the short period natural frequency ($\omega_{SP}$) and damping coefficient ($\zeta_{SP}$).
			   
			\subsection{Analytical Predictions} \label{Analytical-Predictions}
			
			Sixty years ago, the models in the model-test-validate cycle were primarily equations, expressed analytically. To reduce the complexity of these expressions and permit direct solution, simplifying assumptions had to be applied. Typically, this meant setting some parameters to zero or near-constant values. To replicate the predictions of these simplified models, pilots flew maneuvers that achieved these zero or constant values during the maneuver.
						
			This analytical methodology can output a prediction of the short period frequency and damping. To estimate $\bar{M}$ in equation \ref{Mbar} we can construct a skeletal, 2-D model of an aircraft in which a NACA airfoil for the wing is connected by a perfectly rigid, mass-less member to another NACA airfoil for the horizontal tail. 
			
			\begin{wrapfigure}{R}{0.6\textwidth}
				\centering
				\includegraphics[width=0.6\textwidth]{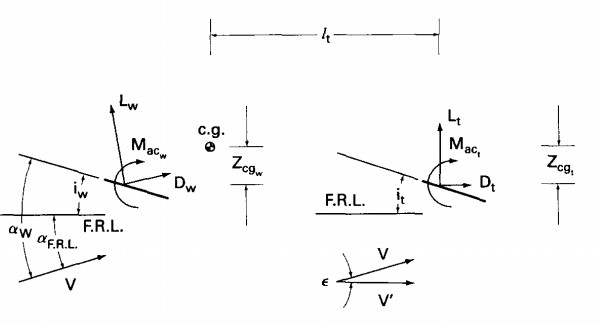}
				\caption{Summing wing and tail moments for a simple airplane model} \label{Moments}
				\vspace{-10pt}
			\end{wrapfigure}
			
			Nelson in fact constructs this model in his monograph on stability and control his sketch of the parameters of interest appears in figure \ref{Moments} \cite{nelson1998stability}. The sum of the moments in this simplified case is simply the moment due to the wing plus the moment due to the tail, each computed between the airfoil aerodynamic center and the aircraft center of gravity. \textit{i.e.},
			
			\[\bar{M} = M_{ac_{w}} + M_{ac_{t}} \]

            Each individual moment can be calculated using the NACA curves for the lift coefficients $C_L$ of the airfoils and the geometry of the aircraft. Since $C_L$ for each airfoil depends on the angle of attack, $\alpha$, then so does the sum of moments. \emph{i.e.}, $\bar{M} = \mathcal{F}(\alpha)$.\footnote{Even in this stick-and-airfoil model the situation is more complicated than this, as the wing exerts a down-wash on the flow that changes the incidence angle of the horizontal tail. The details are unnecessary for the current discussion, and the full treatment can be found in Nelson \cite{nelson1998stability}.} The tail moment must also be a function of the elevator deflection, $\delta_{e}$, since that deflection changes the incidence angle of the tail directly. Less obviously, these moments are also functions of the AOA rate, $\dot{\alpha}$. Combining,

            \begin{equation}
                \bar{M} = \mathcal{F}(\alpha, \dot{\alpha}, \delta_{e})
            \end{equation}

            Given this functional dependency, we can express the sum of moments as a first-order multivariate Taylor series expansion:\footnote{For reasons of notation and unit management we have defined here $M \equiv \frac{\bar{M}}{I_{y}}$}
            
			\begin{equation} \label{TaylorSeries}
				M = \frac{\partial M}{\partial \alpha}\alpha + \frac{\partial M}{\partial \dot{\alpha}}\dot{\alpha} +
				\frac{\partial M}{\partial \delta_{e}}\delta_{e}
			\end{equation}
            
   	        In appendix \ref{1-DOF}, we perform a perturbation analysis on the single degree of freedom (1-DOF) represented by equation \ref{Mbar}. This perturbation analysis studies small angular deviations from steady, trimmed flight, along with other simplifying assumptions. In this analysis motion variables such as $Q$ become perturbative values like $q$, represented by lower-case letters. This allows us to combine equation \ref{Mbar} with equation \ref{TaylorSeries}, producing a second-order ordinary differential equation (ODE) in $\alpha$. In this ODE, terms such as $M_{\alpha}$ are called "stability derivatives," and the subscript indicates partial differentiation by that parameter:

            \begin{equation} \label{ODE-AOA}
                \ddot{\alpha} = M_{\alpha}\alpha + M_{\dot{\alpha}}\dot{\alpha} + M_{\delta_{e}}\delta_{e} 
            \end{equation}         
            
            Since the short period mode is a response to elevator input, we make the time history of the elevator deflection $\delta_{e}(t)$ a forcing function and take the Laplace Transform of both sides. Rearranging:

            \begin{equation} \label{Laplace}
                \frac{\alpha(s)}{\delta_{e}(s)} = \frac{M_{\delta_{e}}}{(s^2 - M_{\dot{\alpha}}s - M_{\alpha})}  
            \end{equation}

            The denominator on the right hand side of equation \ref{Laplace} is called the "characteristic equation" and its roots determine the dynamical behavior of the aircraft angle of attack in response to elevator inputs, which is the definition of the short period. Comparing this denominator to the canonical form of the characteristic equation, $s^2 + 2 \zeta \omega s + \omega^2$, we obtain by inspection:\footnote{The familiar reader will object to the omission of the pitch rate damping term, $M_{q}$, in our expression for $\zeta_{SP}$. We have assumed that term is captured within $M_{\dot{\alpha}}$ for simplicity and readability of this section. }

            \begin{align} 
                \omega_{SP} &= \sqrt{-M_{\alpha}} \label{omega-analytical} \\
                \zeta_{SP} &= \frac{-M_{\dot{\alpha}}}{2 \sqrt{-M_{\alpha}}} \label{zeta-analytical}
            \end{align} 
    
            These allow us to predict the dynamics of the short period mode. Our stick-and-airfoil model allows us to calculate the moments, $M$, at every $\alpha$ of interest, yielding a function $M(\alpha)$. Differentiation of that function supplies $M_\alpha$, which allows us to predict the short period frequency. 
            
            Unfortunately, this is as far as our analytical model takes us. Our simplistic treatment of airflow across airfoils does not yield up a prediction for $M_{\dot{\alpha}}$.\footnote{Although Nelson does provide an analytical expression for this term, under yet more simplifying assumptions. See \cite{nelson1998stability}.} This means we can not predict the short period damping directly with this model. 
            
            Observe also that to arrive at this partial, simplistic prediction of the aircraft's dominant longitudinal mode, we applied a staggering catalogue of simplifying assumptions -- some of which appear in appendix \ref{1-DOF} and others that were omitted for concision. Additionally, each prediction we make is only valid for a particular dynamic pressure, Mach, AOA, and elevator deflection (among others). For a test pilot to validate each prediction, a maneuver must be performed precisely at those parameters and the pilot must faithfully reproduce the assumptions of the foregoing analysis. Any deviations will reduce the accuracy of the results. 
            
            Thus, in this simple example, we have uncovered the origins of databands, tolerances, and the mythos of the ultra-precise flying of the test pilot under extreme conditions. 
			
			\subsection{Computational Predictions} \label{Computational-Predictions}
			
			Today, computers allow for the direct solution of governing equations at a higher level of abstraction, enabling the development of high-fidelity models (HFMs) that address the challenges outlined in the previous section. Despite the additional complexity, these physics-based models still produce physical quantities (such as forces and moments) as outputs. These can be spot-checked through flight test, but only through a series of simplifying assumptions that connect quantities measured during flight test to the underlying physics. 
			
			For example, the field of computational fluid dynamics (CFD) solves the Navier-Stokes equations directly over an entire aircraft with a very high degree of accuracy. The output of a CFD run is a pressure field. This pressure field can be integrated to yield forces and moments such as $\bar{X}$, $\bar{Z}$, and $\bar{M}$ in equations \ref{Xbar} through \ref{Mbar}. Moreover, since CFD can be run at nearly any flow condition, including dynamic conditions with non-zero $q$ and $\dot{\alpha}$, we can use it to calculate a larger number of stability derivatives, including $M_{q}$ and $M_{\dot{\alpha}}$.
   
            This allows us to improve our prediction of the short period mode considerably. In appendix \ref{2-DOF} we show that under a perturbation analysis we can express the EOMs using a larger set of the stability derivatives that we expect to have an impact on system dynamics. The EOMs become:

            \begin{equation}	
        	\begin{bmatrix}
                \dot{u} \\
                \dot{w} - U_{1}q \\
        	    \dot{q} 
            \end{bmatrix} =
        	\begin{bmatrix}
        	  	-g cos \Theta_{1}\theta + X_{u} u + X_{\alpha}\alpha + X_{\delta_{e}}\delta_{e} + X_{\delta_{T}}\delta_{T} \\
        		-g sin \Theta_{1}\theta + Z_{u} u + Z_{\alpha} \alpha + Z_{\dot{\alpha}} \dot{\alpha} + Z_{q} q + Z_{\delta_{e}}\delta_{e} + Z_{\delta_{T}}\delta_{T} \\
        		M_{u} u + M_{\alpha} \alpha + M_{\dot{\alpha}}\dot{\alpha} + M_{q} q + M_{\delta_{e}}\delta_{e} + M_{\delta_{T}}\delta_{T}
        	\end{bmatrix}
        	\end{equation}

            As before, we pose $\delta_{e}(t)$ as the forcing function and take the Laplace transform of both sides. We also assume that $u$ is nearly constant, yielding a 2-DOF approximation. Appendix \ref{2-DOF} shows that this yields a more complex transfer function for $\alpha$:
            
            \begin{align} \label{Computational-TF}
        		\frac{\alpha(s)}{\delta_{e}(s)} &= \frac{ \frac{1}{U_{1}} [Z_{\delta_{e}}s+ (M_{\delta_{e}}U_{1} - M_{q}Z_{\delta_{e}})]}{s^{2} - (M_{q} + \frac{Z_{\alpha}}{U_{1}} + M_{\dot{\alpha}})s + (\frac{Z_{\alpha}M_{q}}{U_{1}} - M_{\alpha})}	
        	\end{align}

            Comparing the denominator of equation \ref{Computational-TF} to the canonical form of a characteristic equation we obtain by inspection
 
            \begin{align} 
		          \omega_{SP} &\approx \sqrt{\frac{Z_{\alpha} M_{q}}{U_{1}} - M_{\alpha}} \label{omega-computational} \\
		          \zeta_{SP} &\approx \frac{-1}{2 \omega_{SP}}(M_{q} + \frac{Z_{\alpha}}{U_{1}} + M_{\dot{\alpha}}) \label{zeta-computational}
	        \end{align}
            
            The higher fidelity of CFD allows us to investigate all of the terms in these approximations, including the previously elusive $M_{\dot{\alpha}}$ and the new $M_q$. This achieves higher accuracy and allows prediction of the dynamics of the aircraft response subject to a wider range parameters of interest to the pilot.
   
           Nevertheless, here we encounter additional difficulties of simplification that constrain the efficacy of flight test. 
           
           Equations \ref{omega-computational} and \ref{zeta-computational} require the partial derivatives $M_{\alpha}$, $Z_{\alpha}$, $M_{q}$, and $M_{\dot{\alpha}}$. We compute those stability derivatives by differentiating the respective functions $M(\alpha), Z(\alpha), M(q),$ and $M(\dot{\alpha})$. Production of these functions requires CFD runs across a very fine grid of $\alpha$, $q$, and $\dot{\alpha}$ values to achieve sufficiently smooth curves, and therefore accurate gradients. As we shall show empirically in section \ref{results}, the short period also varies as a function of dynamic pressure, Mach, and CG location. So these CFD runs must be repeated on another grid of dynamic pressure, Mach, and CG location. CFD models are computationally expensive, so running them across this hyperrectangle of parameters can take hours to days, even on high performance supercomputers. 

           Since the outcome of test is a spot-check of this model at a point, validation of this short period model across an envelope requires the test pilot to fly a sufficient number of validating points across this hyperrectangle. At each point, the pilot must reproduce the parameters of the individual CFD run faithfully, while simultaneously reproducing the assumptions made in appendix \ref{2-DOF} and the foregoing.

            In fact, we should pause here to reflect on a troubling implication of this accumulation of assumptions. It suggests that \emph{there is no such thing as the short period.} Of course, the aircraft AOA responds in some way to elevator input. However, to speak of a short period frequency and damping implicitly accepts the uncoupled, linearized, perturbative, small-angle assumptions that led to equation \ref{Computational-TF}. What the pilot experiences by applying some $\delta_{e}(t)$ to the aircraft system is much more complicated than equation \ref{Computational-TF}. It is, in fact, irreducibly complex. What we call the "short period" is a mathematical metaphor for the dominant elements of the $\frac{\alpha(t)}{\delta_{e}(t)}$ response. It does not appear naked in the physical world, uncoupled from nonlinear dynamics.
            
            This insight will be important as we seek to produce a data-based architecture that prioritizes the data from the short period response over the mathematics of the short period model. The data-based model's prediction of a short period frequency and damping is, in fact, a condescension. The data itself contains richer information than a 2-DOF transfer function. The latter is merely a heuristic that emerges from a mathematical simplification of longitudinal dynamics upon which we have historically relied. 
			
			To summarize: the high accuracy of CFD models allows accurate (but expensive) predictions of functions such as $M(\alpha), Z(\alpha), M(q),$ and $M(\dot{\alpha})$. Differentiation of these functions yields a prediction of the short period frequency and damping through equations \ref{omega-computational} and \ref{zeta-computational}, local to some dynamic pressure, Mach and CG location. To complete the model-test-validate cycle, a test pilot will fly a maneuver at these exact same dynamic pressure, Mach, and CG conditions, which spot-checks the model. If the test data match the predictions, we say the model is “valid” at that point. We do not interrogate the underlying physics of the model, which lie at a level of complexity inaccessible to the test data themselves.
			
			The accuracy of computational HFMs has increased the accuracy of predictions while leaving the primacy of the model intact.
			
			\section{A Data-Based Architecture} \label{DBA}
			
			We propose to invert the relationship between data and model. The data are truth, and the activity of flight test produces these data. These data are used to refine and update a model, which explains the system using physics.
			
			This proposal does not alter or break the model-test-validate cycle. In fact, it strengthens it. The model still comes first chronologically. Test follows. However, the outputs of test are used to refine and update the model in near real-time using contemporary data-based methods. Validation of the model then occurs as a natural consequence of (and simultaneous with) model refinement.  
			
			This architecture reverses the two important consequences of section \ref{MTV}. Since data are collected at any combination of parameters they are in this sense arbitrary. Databands and tolerances do not constrain the test pilot’s airborne execution. More fundamentally, the products of test – the data – immediately and causally update a model, which provides a robust physics-based characterization of the phenomena of interest. This transforms the output of flight test from a spot-check into a direct representation of the underlying physics of the system under test. 
			
			Implementation of this architecture mathematically requires two processes: a forward processes, solved “offline” using high performance compute prior to test; and an inverse process, solved “online” in near real-time from the control room or onboard the aircraft during flight test. 
			
			\subsection{The Forward Process}

\begin{figure}
                \centering
                \includegraphics[width = \linewidth]{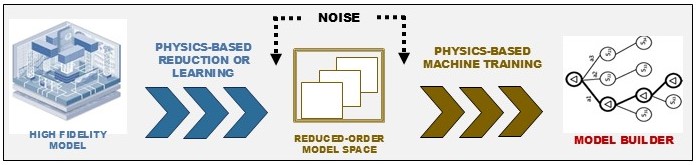}
                \caption{The forward process in a data-based architecture for flight test}
                \label{fig:forward}
            \end{figure}

            The forward process is depicted in figure \ref{fig:forward} and starts with a high fidelity, physics-based digital model of the aircraft under test. For clarity, we refer here to this HFM as though it represented a single flight test discipline, but the overall architecture is extensible to multidisciplinary envelope expansion by coupling multiple HFMs. 
            
            From the HFM, reduced-order models (ROMs) are developed. These require less computational resources, but are not merely correlational heuristics; rather, they remain physics-based and traceable to the high-fidelity models. In section \ref{flight-test} we will present a simple example of physics-based reduction used in a flight test campaign. Section \ref{future} will then suggest more generalizable and more accurate methods. 
            
            After instantiating some reduced-order model, a "model space" of ROMs is created. We may think of this ROM space as containing  many slightly different, discrete versions of the ROM. In each version, the parameters that particularize the model vary, consistent with their uncertainty bounds. The structure or structures of the ROMs, along with all permutations of physically plausible parameter values comprise the entire model space. 
            
            The final step of the forward process is to train a model builder from the ROM space, using physics-constrained, data-based methods. This model builder will comprise the algorithmic heart of the inverse problem.

            On first encounter, it can be useful to conceive of the construction of a model builder from a ROM space using a pure imitation-learning (IL) metaphor. If, subject to some set of inputs, each ROM outputs a physical quantity of interest that characterizes the aircraft under test, then we can run each ROM subject to thousands -- or millions -- of the possible inputs. We use the input-output space to train a secondary algorithm (the model builder) to associate each input-output mapping with a particular ROM, using some computationally sparse method. In flight test, when the model builder observes some actual input-output pair, it then "selects" the ROM most likely to have yielded that mapping. For this reason, a pure IL model builder is sometimes called a model selector.

            However, we emphasize that pure IL as we have just described is a poor end state for this architecture. An IL model selector is free to "learn" erroneous mappings, particularly in corner cases, and make non-physical ROM selections when dealing with multiple disciplines. In our architecture, we insist that physics be enforced as much as possible and suggest methods for doing so in section \ref{future}.

            A chief difficulty of flight test results from the noisiness of the data. For that reason, our architecture also permits the insertion of noise in two separate places, as illustrated in figure \ref{fig:forward}. First, noise may be inserted during the construction of the ROM space itself. This multiplies the size of the ROM space, because now each ROM exists in multiple versions, each permuted by some physically plausible noisiness of the parameters that particularize the model. Second, noise may be inserted during the training of the model builder. In this case, it is the inputs and outputs that are distorted by a physically plausible noise distribution. The model builder then must learn this noisy input-output mapping. 

            The end product of the forward process is the model builder, which embeds knowledge of the ROM space into some trained mapping, while enforcing physics both locally and globally. Whatever data-based training technique is used, the builder effectively maximizes the probability that an observed input-output pair would result from the form of the constructed (or selected) reduced-order model. 

            \subsection{The Inverse Process}
            
\begin{figure}
                \centering
                \includegraphics[width = \linewidth]{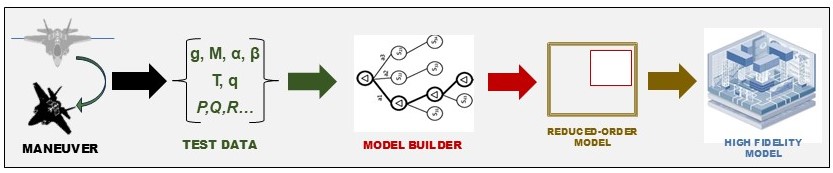}
                \caption{The inverse process in a data-based architecture for flight test}
                \label{fig:inverse}
            \end{figure}

            With a trained model builder, flight testers can now accomplish the inverse process, as depicted in figure \ref{fig:inverse}. During flight test, an air vehicle executes some arbitrary maneuver, which generates time-varying data collected by a suite of sensors.\footnote{As we have qualified above, this maneuver is only "arbitrary" in the sense that the test pilot is not limited by databands and tolerances.The test pilot remains limited by the need to execute a build-up approach within an envelope, by the need to test at sufficient granularity within the envelope, the need to gather information about the relevant physics, and all the constraints related to risk management and statistical sufficiency.}

            These data are also arbitrary, in the sense that no unyielding requirement exists to have certain parameters (e.g. acceleration) driven to zero and other parameters (e.g. Mach) maintained at constant values. It certainly may be the case that parametric constraints such as these do exist as part of the maneuver the test pilot executes. However, they exist as guidelines and not mandates. The model builder can operate usefully from the data even if all such parametric guidelines are violated.
            
            When the trained model selector receives the flight test data, it builds (or selects) the most probable model that would have produced the observed data. This process occurs in "near real-time," by which we mean we intend to execute it after every single maneuver in a flight test sortie. This means it should require at most the time it takes the pilot to set up for a subsequent maneuver -- a few minutes, usually. 
            
            We demand near real-time performance partly because after a single maneuver, the uncertainty that the constructed model is the correct one will be high. After a second maneuver, an entirely different ROM might result. However, as massive quantities of data are ingested, ordinary statistical methods may be utilized to gain confidence in a particular region of the ROM space. That particular region of the ROM space we then accept as the region of actual air vehicle performance. 
            
            The final step in the data-based architecture is to update the high-fidelity digital models using the parameters identified in the reduced order models. In general this is a manual process. In some cases, it may be automated. However the feasibility of automating the traceability between a high-fidelity model and a reduced-order model depends on the format of the high-fidelity model and the technique chosen for reducing the order of the model. 

            \subsection{Advantages of a Data-Based Architecture} \label{advantages}

            The data-based architecture presented in this section has many advantages over the model-based methods currently used across the Department of Defense. We have suggested many of them in the foregoing, but reprise them here for completeness.
    
            \emph{\textbf{The Elimination of Test Points}}

            Under this architecture, data need never be discarded for the test pilot falling out of a databand or tolerance. The model is refined \textit{from} the data, so any data will do. This (and only this) is what we mean when we refer to the methods of this architecture as "Pointless."
            
            The analysis presented in section \ref{Motivation} indicates that this one single advantage would have accelerated system design and development of a major DoD program by a factor of six. 

            \emph{\textbf{The Representation of Physics from Test}}

            The function of flight test is to collect data. Under this data-based architecture, the data run a model builder, which updates a ROM. But that ROM, being a model, is a physics-based representation of the phenomenon under study. It can interpolate, it can extrapolate, and it can explain.
            
            Consider the case study of section \ref{Computational-Predictions}. That example showed that CFD could be used to generate a point prediction of $\omega_{SP}$ and $\zeta_{SP}$ using partial differentiation of the functions $M(\alpha), Z(\alpha), M(q),$ and $M(\dot{\alpha})$. But if we directly measure $\omega_{SP}$ and $\zeta_{SP}$ during flight test and not the functions themselves, the outcome of flight test gives us no access to the embedded physics. In fact, we can not even convert our measurement of $\omega_{SP}$ to a point measurement of $M_{\alpha}$, because through equation \ref{omega-computational} it is confounded with $Z_{\alpha}$ and $M_{q}$!

            By contrast, the data-based architecture ingests the same data that lead to a point prediction of $\omega_{SP}$ and $\zeta_{SP}$. However, it returns a ROM. That ROM has embedded within it a representation of the functions $M(\alpha), Z(\alpha), M(q),$ and $M(\dot{\alpha})$. This allows us as a very trivial matter to reconstruct the values of all the stability derivatives associated with those functions. 
            
            Our architecture shares this advantage with other types of system identification (SysID), which also produce parameterized ROMs. However, the forms of time-domain and frequency domain SysID commonly implemented in flight test constrain the state variables and the structure of the identified system \cite{morelli2023sysid}. Our architecture incorporates and generalizes SysID for model building; it admits a completely general state space and arbitrarily non-linear ROMs.\footnote{The authors continue to advocate for the implementation of SysID methods over traditional flight test techniques in envelope expansion.}
          
            We shall show in section \ref{results} that the physical representation our architecture outputs allows us to generate physics-based predictions about parameters external to our mathematical representation of some phenomenon. As an example, we learned $\omega_{SP}$ for the T-38 from flight test data. Dynamic pressure, $\bar{q}$ never appears in sections \ref{Analytical-Predictions} or \ref{Computational-Predictions}, but $\omega_{SP}$ depends on it implicitly. Because of the implicit dependence, our test data allowed us to reconstruct a functional dependence $\omega_{SP} = \mathcal{F}(\bar{q})$ that accurately predicted the short period at dynamic pressures we never tested. 

            \emph{\textbf{The Quantification of Uncertainty}}

            Consider a simple flight test technique that yields a single measurement of some quantity of interest. After making such a measurement, the test team can say very little about its confidence in the measured value. To gain such confidence, the team sends out the pilot to repeat the measurement. It must be repeated a number of times, dependent on the size of the measurement relative to the sources of error, to quantify the level of uncertainty associated with the set of measurements.

            By contrast, \textit{certain} data-based methods that train a model builder on a ROM space can quantify uncertainty on the constructed or refined model with a single measurement. This is because sources of noise and uncertainty are characterized by error distributions during the training of the model itself. In section \ref{gpr-application}, we show an example of a model builder using Gaussian Process Regression, which explicitly quantifies uncertainty at every point on its ROM, and updates that uncertainty after every flight test measurement.

            This does not suggest that uncertainty is spontaneously reduced using this architecture. Rather that we get a quantification of that uncertainty earlier, by incorporating priors about the distributions of error sources.

            \emph{\textbf{The Outcome-based Framework for Risk Management}}

            In their presentation to the Society of Experimental Test Pilots in 2022, Jurado and McGehee proposed a novel "outcome-based framework" for risk awareness \cite{jurado2022risk}. Their paper "operationalized" the risk-awareness model proposed by Wickert in 2018 by replacing existing flight test continuation criteria (FTCC) with "knowledge envelopes \cite{wickert2018risk}."
            
            A knowledge envelope is constructed mathematically by permuting some functional representation of some measurable quantity of interest many times. Each permutation slightly changes the parameters that govern the shape of the function, within the range of possible parameter values. This results in a family of functions, all possible descriptions of the actual aircraft behavior. Each function is then labeled as "safe" or "unsafe;" the boundary between safe and unsafe parameter permutations delimits the knowledge envelope. 
            
            This process of permuting functional representations of aircraft quantities of interest is identical to our process for creating a ROM space, where ROMs take the place of functional representations of the system. In fact, Jurado and McGehee refer to these representations as "aircraft models." Our ROM space generalizes these functional models, because each ROM may contain many functions. The complete set of functions in a ROM fully describes the flight test phenomenon under study. 
            
            By implication, the data-based architecture presented in this paper immediately enables the Jurado-McGehee outcome-based risk management framework. After building a ROM space, we classify each ROM as "safe" or "unsafe" according to some physics-based criteria. We then construct a knowledge envelope (which in general is of high dimension) from the parameters that particularize the ROMs.\footnote{Our architecture also supports multivariate labeling of safe-unsafe within the ROM. That is to say, some aircraft parameters can be safe while others are unsafe. Any unsafe parameter will lead to an unsafe classification overall. However, this multivariate labeling yields higher fidelity insight into the reason for the knowledge envelope excursion. Additionally, we can use continuous rather than binary scales of safeness.} We use the knowledge envelope to manage risk during envelope expansion.

            \section{Flight Test Application} \label{flight-test}

            We conducted a short flight test campaign on the T-38C as a staff Test Management Program (TMP) at Test Pilot School (TPS). This campaign exercised a rudimentary version of the data-based architecture proposed in section \ref{DBA}. The following presentation of the methodology and results should thus be viewed as a proof-of-concept demonstration of our architecture. We do not propose to use the specific mathematical methods used here for real-world envelope expansion programs. We present specific methods that we do believe are extensible to DoD flight test programs in section \ref{future}.


            \subsection{Methodology}

\begin{figure}
                \centering
                \includegraphics[width = \linewidth]{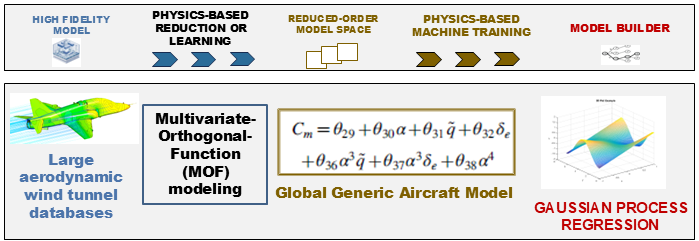}
                \caption{The forward process as implemented for a T-38 flight test campaign}
                \label{fig:forward-ex}
            \end{figure}

            Figure \ref{fig:forward-ex} depicts the implementation of the forward process in our flight test campaign, juxtaposed below the generic architecture from figure \ref{fig:forward}. The compliance with the architecture is imperfect; however the fundamental inversion of the relationship between data and model was achieved.
            
            For the data-based reduction of a high-fidelity model to a ROM space, we implemented the Global Generic Aerodynamic Model developed by Grauer and Morelli \cite{morelli2015global}. That model begins with a large database of aerodynamic coefficients collected in a wind tunnel using both static and forced-oscillation investigations. This database, which represented the physics of the aircraft to a very high degree of fidelity, acted as our HFM.

            To this high fidelity database, a model order reduction technique known as Multivariate Orthogonal Function (MOF) modeling was applied. This technique is beyond the scope of this paper, but its elaboration can be found in the references \cite{morelli1995mof}, \cite{morelli1998f16}, and \cite{morelli2006sysid}.

            The process of MOF reduced the high-order databases to a linear model that depended on forty different model parameters, $\{\theta_{i}\}$. These parameters were coefficients on model variables identified through a model structure determination protocol. For example, the pitching moment coefficient was expressed as

            \begin{equation} \label{mean-function}
                C_{m} \equiv \frac{M}{\bar{q}S\bar{c}} = 
                \theta_{29} + \theta_{30} \alpha + \theta_{31} \tilde{q} + \theta_{32} \delta_e + \theta_{33} \alpha \tilde{q} + \theta_{34} \alpha^2 \tilde{q} + \theta_{35} \alpha^2 \delta_e + \theta_{36} \alpha^3 \tilde{q} + \theta_{37} \alpha^3 \delta_e + \theta_{38} \alpha^4 
            \end{equation}

            where $\tilde{q}$ is the non-dimensional pitch rate. Compare this generic expression for $M$ with equation \ref{TaylorSeries}. Evidently, $\theta_{30}$ is proportional to $\frac{\partial M}{\partial \alpha}$. However, higher-order dependencies up to $\alpha^4$ are captured in equation \ref{mean-function}.

            This linear model spans a ROM space, since each coefficient can take any real value.\footnote{This is a good example of a ROM space that does not require the explicit permutation and storage of the parameters of an individual
            ROM. By accepting any combination of real-valued parameters, the ROM expressed by equation \ref{mean-function} contains an infinite number of permutations.} However, we did not computationally permute the expression for the ROM. Instead, we adopted a particular ROM, with particular model parameters $\{\theta_{i}\}$ as a mean function for the model builder. In fact, we adopted the particular model of the A-7E Corsair II.
            
            \begin{wrapfigure}{R}{0.4\linewidth}
                \centering
                \includegraphics[width = 0.8\linewidth]{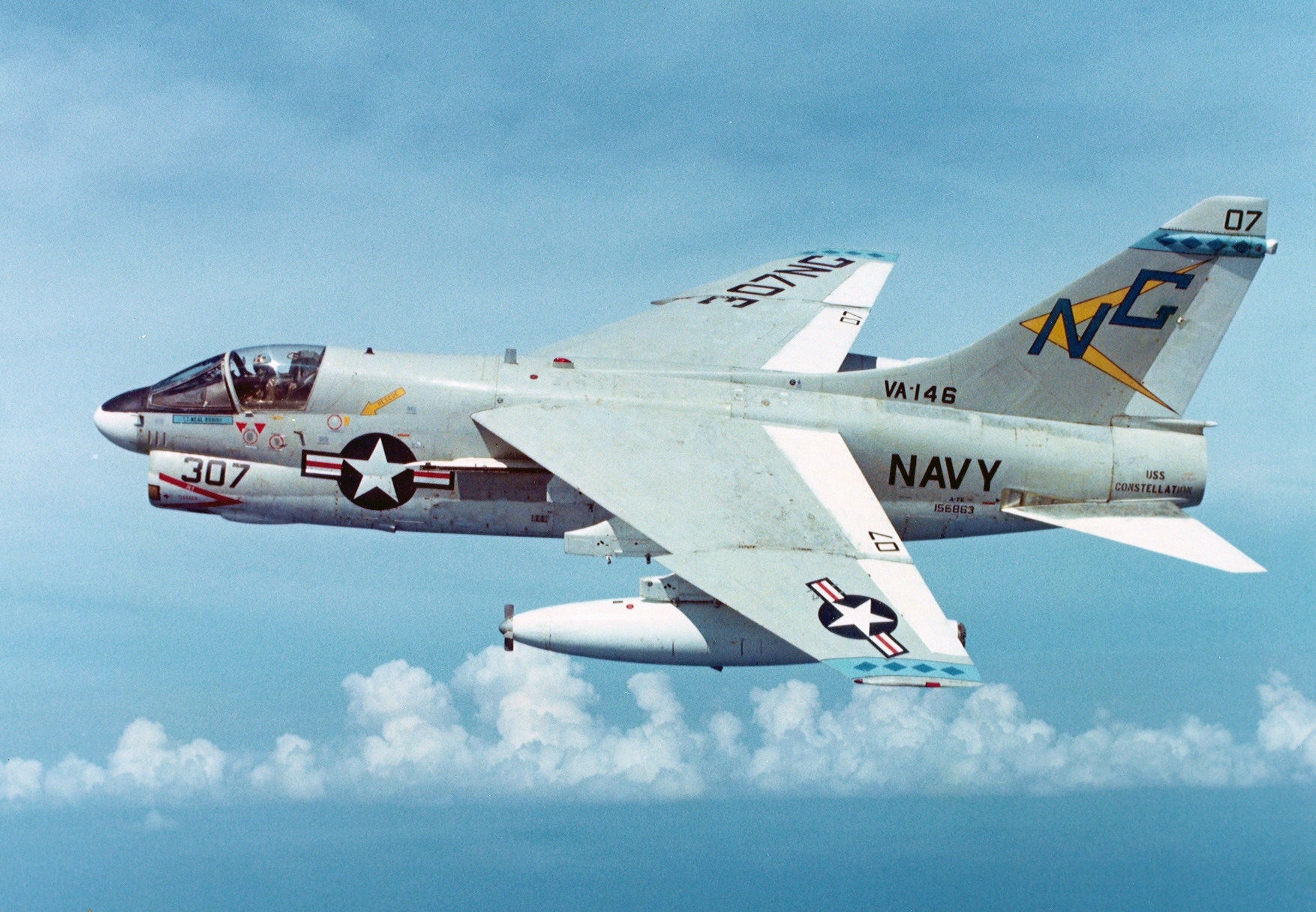}
                \caption{The A-7E Corsair II, the aerodynamics of which were used as a mean function for regressing flight test data for the T-38C.}
                \label{fig:A7}
            \end{wrapfigure}
            
            The mathematics of the mean function will be explained more fully in section \ref{gpr-application}. However conceptually we adopted the particular ROM of the A-7 as our initial mean function for the T-38 for two reasons. First, it embedded physics into our model builder, which is otherwise a na\"ive machine-learning technique. Second, we intentionally used a version of the ROM that was known to be incorrect. Photographed in figure \ref{fig:A7}, the A-7 was manufactured four years after the T-38. While of a similar generation of aircraft, it was substantially larger than the T-38 and differed in stability and control characteristics. We found this desirable for our experiment; we wanted to start with a similar but fundamentally erroneous ROM, and have the data-based method resolve the discrepancies.

            With this generic aerodynamic model of the A-7 as our ROM, we selected Gaussian Process Regression (GPR) as our model builder algorithm for the inverse problem. The theory of GPR as applied in this case will be elaborated in section \ref{GPR}. For our current purposes, GPR takes the initial mean function (in this case, the A-7) and then ingests test data to re-shape that function into a mathematical object that embeds the observed flying qualities of the airplane. Having selected GPR, we concluded in implementing the forward process.   

\begin{figure}
                \centering
                \includegraphics[width = \linewidth]{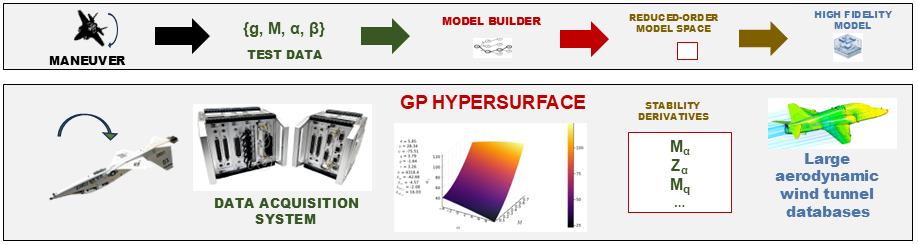}
                \caption{The inverse process as implemented for a T-38 flight test campaign}
                \label{fig:inverse-ex}
            \end{figure}

            The implementation of the inverse process is depicted in figure \ref{fig:inverse-ex}. A single T-38 sortie was flown to gather data. On that sortie, \emph{rollercoaster} maneuvers were flown at dynamic pressures that matched historical estimates of the T-38 short period and damping. In a \emph{rollercoaster} maneuver, the pilot began by trimming the aircraft hands-off in wings-level, unaccelerated flight at a particular altitude and Mach.\footnote{The data from these trim points were independently required to carry out our analysis, as will be explained more fully in section \ref{gpr-sp}.} Then, the pilot smoothly oscillated the aircraft in pitch between zero and two Gs at a rate of approximately 0.5 G per second, while maintaining wings level. No other datbands or tolerances were applied to the maneuver, other than targeting particular Mach and altitude combinations for reasons of historical comparison.

            We chose the \emph{rollercoaster} maneuver not because it was necessary, but simply as an easily understood method to gather dynamic data in the vicinity of 1.0 G. Additionally, we specifically \textit{excluded} flight test techniques historically used to measure the short period (steps, doublets, frequency sweeps), in order to test our claim that arbitrary data could be used to make predictions. 
            
            Full aircraft state data, including calibrated control surface deflections, were recorded onto a data acquisition system (DAS) during each maneuver. Because the data were recorded and not telemetered, model building was not performed during the flight test. Instead, the model builder used the data after the sortie to perform GPR, using our ROM as the mean function. We did not formally assess run-time; however in general we could construct a hypersurface from a one-minute maneuver in about one second. This exceeded our criterion for near real-time operation.

            As we shall see in section \ref{gpr-application}, the output of GPR is an infinitely-differentiable hypersurface with an uncertainty associated with each point on the hypersurface. Because we could differentiate the hypersurface along any dimension, we were able to calculate all the stability derivatives of interest. These stability derivatives were comparable to the $\theta$ values that appear in equation \ref{mean-function} (except that the structure of our model was allowed to be highly non-linear). Because of that, these derivatives comprised a new form of a particular ROM, which we identified as the physics of our system for longitudinal flying qualities.

            The final step of implementing the inverse process should be to update the aerodynamic databases (the HFM) with values obtained during flight test. We did not accomplish this, as our aerodynamic database was for the A-7 not the T-38. To reiterate, this discrepancy was intentional.
          
            \subsection{Gaussian Process Regression as Model Builder} \label{GPR}

            In the remainder of this section, we demonstrate how Gaussian Process Regression (GPR) builds a model of the aircraft's flying qualities by learning from test data. This allows us to extract a ROM of stability derivatives directly from the hypersurface created by the Gaussian process. We then use this ROM to compute key dynamic characteristics such as short period frequency and damping, comparing the results to previous findings.

            \subsubsection{GPR Theory} The following uses Gaussian process notation consistent with that of \citeauthor{kochenderfer2019algorithms} \cite{kochenderfer2019algorithms}. A Gaussian process defines a distribution over functions. Using Gaussian processes we can model noisy observations $y$ of the true function $f$ that we are interested in predicting. These noisy observations are given by $y = f(\mathbf{x}) + z$ where $f$ is deterministic but $z$ is zero-mean Gaussian noise, $z \sim \mathcal{N}(0, \nu)$. If we already have a set of observed points $X$ and the corresponding $\mathbf{y}$, we can predict the values $\hat{y}$ at points $X^*$. The joint distribution is given by: \begin{equation}
            \begin{bmatrix}
                \hat{y} \\
                y
            \end{bmatrix}
            \sim \mathcal{N} \left( 
                \begin{bmatrix}
                \mathbf{m}(X^*) \\
                \mathbf{m}(X)
                \end{bmatrix}, 
                \begin{bmatrix}
                \mathbf{K}(X^*, X^*) & \mathbf{K}(X^*, X) \\
                \mathbf{K}(X, X^*) & \mathbf{K}(X, X) + \nu \mathbf{I}
                \end{bmatrix} 
            \right)
            \end{equation} with conditional distribution $\hat{y} \mid y, \nu \sim \mathcal{N}(\mu^*, \Sigma^*)$ where: \begin{equation}
            \begin{aligned}
                \mu^* &= \mathbf{m}(X^*) + \mathbf{K}(X^*, X)(\mathbf{K}(X, X) + \nu \mathbf{I})^{-1}(y - \mathbf{m}(X)) \\
                \Sigma^* &= \mathbf{K}(X^*, X^*) - \mathbf{K}(X^*, X)(\mathbf{K}(X, X) + \nu \mathbf{I})^{-1} \mathbf{K}(X, X^*).
            \end{aligned}
            \end{equation} In the equations above, we use the functions $\mathbf{m}$ and $\mathbf{K}$ which are given by: \begin{equation}
                \begin{aligned}
                    \mathbf{m}(X) &= \begin{bmatrix} m(x^{(1)}), \dots, m(x^{(n)}) \end{bmatrix} \\
                    \mathbf{K}(X, X') &= 
                    \begin{bmatrix}
                    k(x^{(1)}, x'^{(1)}) & \dots & k(x^{(1)}, x'^{(m)}) \\
                    \vdots & \ddots & \vdots \\
                    k(x^{(n)}, x'^{(1)}) & \dots & k(x^{(n)}, x'^{(m)})
                    \end{bmatrix}.
                \end{aligned}
            \end{equation} Here, $m(\mathbf{x})$ represents the \textit{mean function}, while $k(\mathbf{x}, \mathbf{x'})$ is the \textit{covariance function} or kernel. The mean function encodes prior knowledge about the function’s behavior, and the kernel dictates the smoothness and other properties of the functions.

            \subsubsection{GPR Application} \label{gpr-application}
            In this work, we used GPR to build a probabilistic model that mapped aircraft state variables to aerodynamic quantities of interest, specifically the pitching moment coefficient \( C_m \). The state vector \( \mathbf{x} \) encompassed key flight parameters: \begin{equation}
                \mathbf{x}^{(i)} = \begin{bmatrix}
                    M^{(i)} \\
                    \rho^{(i)} \\
                    \bar{q}^{(i)} \\
                    P^{(i)} \\
                    Q^{(i)} \\
                    R^{(i)} \\
                    \alpha^{(i)} \\
                    \delta_e^{(i)}
                \end{bmatrix} \label{eq:state_vector},
            \end{equation} where \( M \) is the Mach number, \( \rho \) is the air density (pre-computed from pressure altitude and outside air temperature), \( \bar{q} \) is the dynamic pressure, \( P, Q, R \) are the roll, pitch, and yaw rates, \( \alpha \) is the angle of attack, and \( \delta_e \) is the stabilator deflection.\footnote{Observe that we use non-perturbation variables here. In section \ref{differentiation} we will define stability derivatives with respect to some perturbation variable as equivalent to the construction of a linear gradient in the state variable, at some trim point.}
            
            Our objective was to learn the underlying function \( f: \mathbf{x} \mapsto C_m \) that mapped these state variables to the pitching moment coefficient. To achieve this, we constructed a Gaussian process over \( f \) and updated it using observed data \( \{ (\mathbf{x}^{(i)}, y^{(i)}) \}_{i=1}^N \), where \( y^{(i)} \) were noisy observations of \( C_m \).
            
            We incorporated physics-based priors into the mean function \( m(\mathbf{x}) \) of the Gaussian process, reflecting established aerodynamic relationships. Specifically, we used the mean function derived from the Morelli aerodynamic model, expressed in equation \ref{mean-function}. This mean function encoded our prior knowledge about how \( C_m \) depended on \( \alpha \), \( q \), and \( \delta_e \), allowing the Gaussian process to focus on modeling deviations from this baseline.
            
            
            \begin{figure}
                \centering
                \includegraphics[width = 0.8\linewidth]{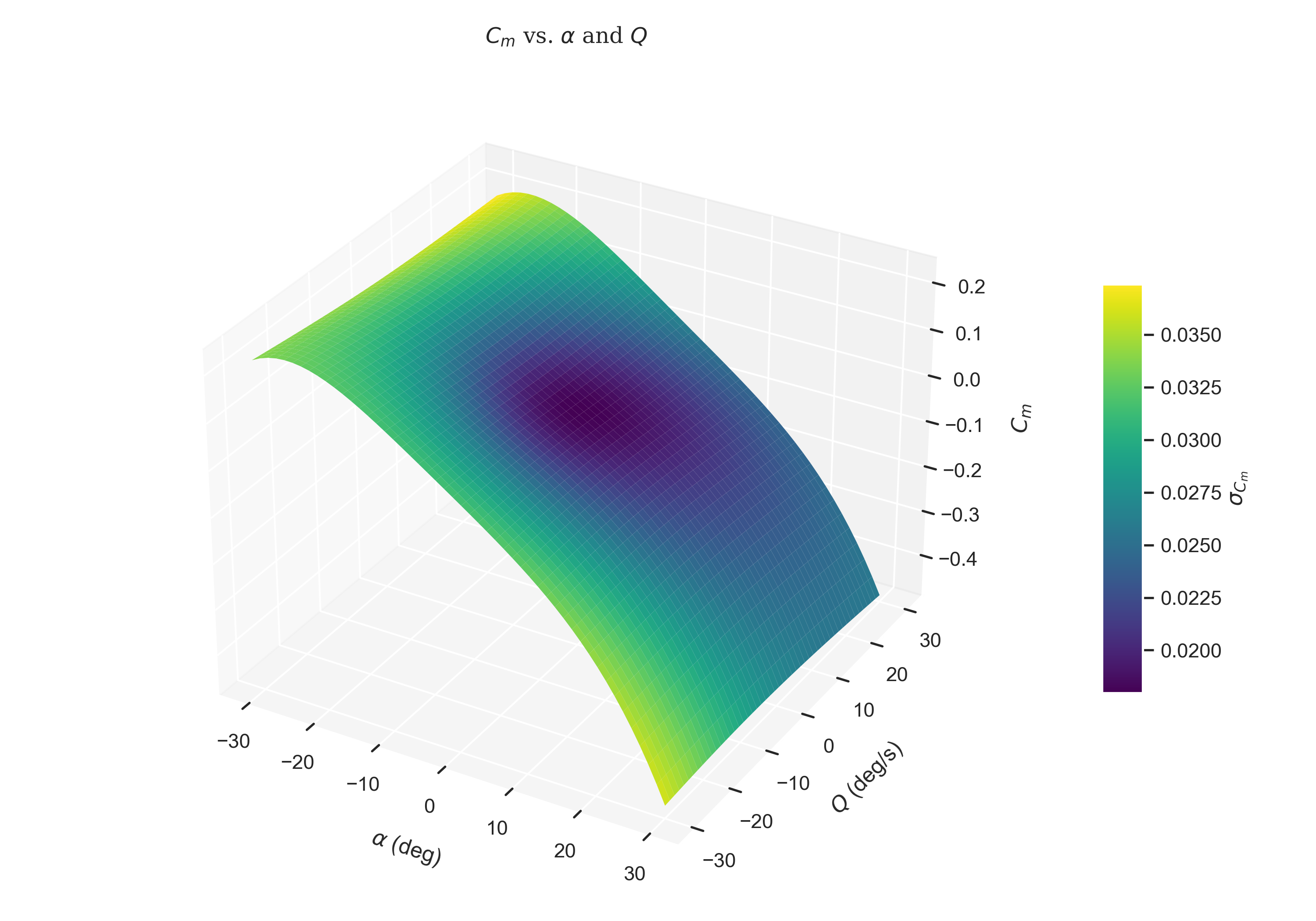}
                \caption{Surface plot showing the relationship between \( \alpha \), \( Q \), and \( C_m \) learned by the Gaussian process. The surface is colored by the model's uncertainty values.}
                \label{fig:surface_plot}
            \end{figure}
            
            The covariance function \( k(\mathbf{x}, \mathbf{x}') \) determined the smoothness and generalization properties of the Gaussian process. We employed the Neural Network kernel:
            
            \begin{equation}
                k(\mathbf{x}, \mathbf{x}') = \sin^{-1} \left( \frac{\mathbf{x}^\top \mathbf{x}'}{\sqrt{(1 + \|\mathbf{x}\|_2^2 / 2)(1 + \|\mathbf{x}'\|_2^2 / 2)}} \right),
            \end{equation} which captures complex, non-linear relationships between inputs and outputs.

            After training, the Gaussian process provides a posterior distribution over functions, enabling us to make predictions about \( C_m \) at new state vectors \( \mathbf{x}^* \). The predictive mean and variance were given by:
            
            \begin{equation}
                \begin{aligned}
                    \mu(\mathbf{x}^*) &= m(\mathbf{x}^*) + \mathbf{k}(\mathbf{x}^*, \mathbf{X})^\top (\mathbf{K}(\mathbf{X}, \mathbf{X}) + \nu \mathbf{I})^{-1} (\mathbf{y} - \mathbf{m}(\mathbf{X})), \\
                    \sigma^2(\mathbf{x}^*) &= k(\mathbf{x}^*, \mathbf{x}^*) - \mathbf{k}(\mathbf{x}^*, \mathbf{X})^\top (\mathbf{K}(\mathbf{X}, \mathbf{X}) + \nu \mathbf{I})^{-1} \mathbf{k}(\mathbf{X}, \mathbf{x}^*),
                \end{aligned}
            \end{equation} where \( \mathbf{k}(\mathbf{x}^*, \mathbf{X}) \) is the covariance vector between the test point and training inputs, and \( \nu \) is the noise variance.
            
            An example of a hypersurface extracted from the Gaussian process is shown in figure \ref{fig:surface_plot}. In this 3-D hypersurface, we have shown how changes in $\alpha$ and $q$ influenced the pitching moment coefficient, by setting all other state variables (the other dimensions of the hypersurface) to some constant values. Observe that the Gaussian process also provides uncertainty estimates; the model is most certain at median values of $\alpha$ and $q$, as we would expect.

            \subsection{Prediction of the Short Period} \label{gpr-sp}
            One of the key advantages of GPR is that it provides a smooth, differentiable hypersurface from the posterior estimate. This allowed us to compute classical stability derivatives by differentiating the Gaussian process at specific flight conditions. 

            \subsubsection{Trim Conditions and Stability Derivatives} \label{trim}

            To analyze the aircraft's dynamic behavior and compare with historical databases, we focused on specific flight configurations known as \emph{trim conditions}. In our usage, a trim condition is a steady-state flight condition where the aircraft was in equilibrium, so that the rate components of the state vector were zero. In order to query the Gaussian process at multiple trim points to compare with historical data, we first needed to determine the trim function for the aircraft. 
            
            We experimentally determined trim functions by performing \emph{trim shots} at different altitudes and airspeeds. During these maneuvers, the aircraft was flown in steady, unaccelerated flight and the corresponding state variables were recorded. After the flight, we regressed the data to obtain empirical relationships for the trim angle of attack \( \alpha_{\text{trim}} \) and the trim stabilator deflection \( \delta_{e_{\text{trim}}} \) as functions of dynamic pressure \( \bar{q} \):
            
            \begin{align}
                \alpha_{\text{trim}}(\bar{q}) &= a \cdot \exp(-b \bar{q}), \label{trim-aoa} \\
                \delta_{e_{\text{trim}}}(\bar{q}) &= c + d \cdot \ln(\bar{q}), \label{trim-stab}
            \end{align} where \( a, b, c, \) and \( d \) were coefficients obtained from the regression of experimental data. To account for Mach effects, we calculated multiple trim functions for multiple Mach ranges.
            
            These relationships allowed us to define the trim state \( \mathbf{x}_{\text{trim}} \) at any given dynamic pressure. Mathematically, we computed the stability derivatives at these trim states in order to compare with historical data.

            \subsubsection{Differentiation of the Gaussian Process} \label{differentiation}
            
            The stability derivatives were obtained by differentiating the Gaussian process posterior mean function \( \mu(\mathbf{x}) \) with respect to the relevant state variables at the trim condition \( \mathbf{x}_{\text{trim}} \). Specifically, we computed:
            
            \begin{align}
                C_{m_\alpha} &\equiv \left. \frac{\partial \mu(\mathbf{x})}{\partial \alpha} \right|_{\mathbf{x} = \mathbf{x}_{\text{trim}}}, \label{Cm_alpha} \\
                C_{m_q} &\equiv \left. \frac{\partial \mu(\mathbf{x})}{\partial Q} \right|_{\mathbf{x} = \mathbf{x}_{\text{trim}}}.  \label{Cm_Q}
            \end{align}
            
            The posterior mean function \( \mu(\mathbf{x}) \) was given by:
            
            \begin{equation}
                \mu(\mathbf{x}) = m(\mathbf{x}) + \mathbf{k}(\mathbf{x}, \mathbf{X})^\top \mathbf{A},
            \end{equation}
            
            with
            
            \begin{equation}
                \mathbf{A} = (\mathbf{K}(\mathbf{X}, \mathbf{X}) + \nu \mathbf{I})^{-1} (\mathbf{y} - \mathbf{m}(\mathbf{X})).
            \end{equation}
            
            The gradient of \( \mu(\mathbf{x}) \) with respect to \( \mathbf{x} \) was:
            
            \begin{equation}
                \nabla_{\mathbf{x}} \mu(\mathbf{x}) = \nabla_{\mathbf{x}} m(\mathbf{x}) + \left( \nabla_{\mathbf{x}} \mathbf{k}(\mathbf{x}, \mathbf{X}) \right)^\top \mathbf{A}. \label{eq:grad_gp_mean}
            \end{equation}
            
            We used automatic differentiation tools to compute the gradients of equation \ref{eq:grad_gp_mean} efficiently. 

            \subsubsection{Calculation of Short Period Dynamics}

            With the stability derivatives \( C_{m_\alpha} \) and \( C_{m_q} \) obtained from the Gaussian process, we calculated the short period natural frequency \( \omega_{n_{sp}} \) and damping ratio \( \zeta_{sp} \), which are useful parameters for understanding the aircraft's longitudinal dynamic response.
            
            The short period dynamics are derived from the linearized longitudinal equations of motion. The expressions for \( \omega_{n_{sp}} \) and \( \zeta_{sp} \) given in \ref{omega-computational} and \ref{zeta-computational}, and reproduced here were:
            
            \begin{align}
                \omega_{SP} &= \sqrt{ \frac{ -Z_{\alpha} M_q }{ U_1 } - M_{\alpha} }, \label{omega-gp} \\
                \zeta_{SP} &= - \frac{ M_q + \left( M_q / 3 \right) + (Z_{\alpha} / U_1) }{ 2 \omega_{SP} }, \label{zeta-gp}
            \end{align} 
            
            
            except in \ref{zeta-gp} we have replaced $M_{\dot{\alpha}}$ with $\frac{1}{3}M_q$ as proposed by \citeauthor{yechout2003introduction} in \cite{yechout2003introduction}.

            \subsection{Results} \label{results}


            To assess the accuracy of our methods, we gathered historical comparison data on the T-38 short period mode from four sources, encompassing a range of methods. 

            \begin{figure}
                \centering
                \includegraphics[width = 0.8\linewidth]{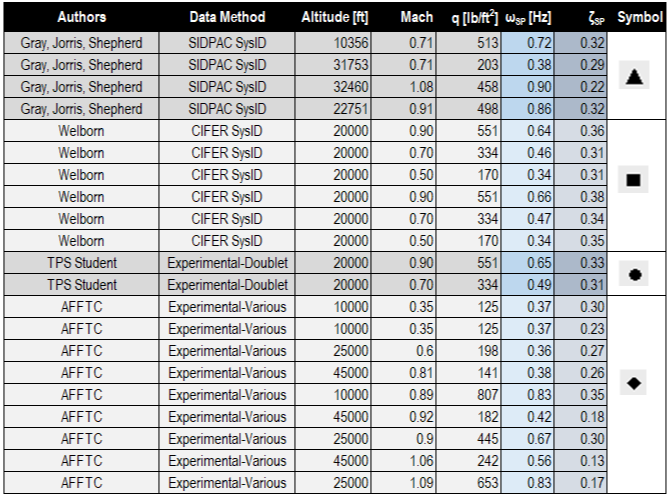}
                \caption{The consolidated short period data collected for comparison. The symbols cross-reference with figure \ref{fig:comparison-points}.}
                \label{fig:SP-table}
            \end{figure}   

            \begin{enumerate}
                \item A system identification (SysID) investigation flown by the Test Pilot School in the T-38A\footnote{We do not expect significant differences in the short period between the T-38A and the T-38C}. This investigation used frequency sweeps to excite the short period mode, and the System Identification Programs for AirCraft (SIDPAC) toolbox for MATLAB to identify stability derivatives using a linear model structure \cite{shepherd2010limited}.
                
                \item A system identification investigation flown by a Master's student for her thesis. This investigation used the Comprehensive Identification from FrEquency Responses (CIFER®) software package to identify transfer functions directly. The results included $\frac{\alpha}{\delta_e}$ transfer functions and $\frac{\theta}{\delta_e}$ transfer functions, both of which were used for comparison \cite{welborn2010tfs}.
                
                \item Two T-38 flying qualities reports completed by TPS students. These investigations both used doublets to excite the short period mode, and the amplitudes and phases of the oscillation were calculated directly from the DAS \cite{havekirby, havepalpatine}.
                
                \item The Air Force Flight Test Center (AFFTC) final stability and control report on the T-38A. Various methods were used, as explained in the reference \cite{dtic1961t38}\footnote{Pitch damper OFF and pitch damper ON stability and control results were presented in the test report. We used the pitch damper OFF values, since the pitch damper was removed for the T-38C \cite{grayemail}.}.
            \end{enumerate}

            \begin{figure}
                \centering
                \includegraphics[width = 0.8\linewidth]{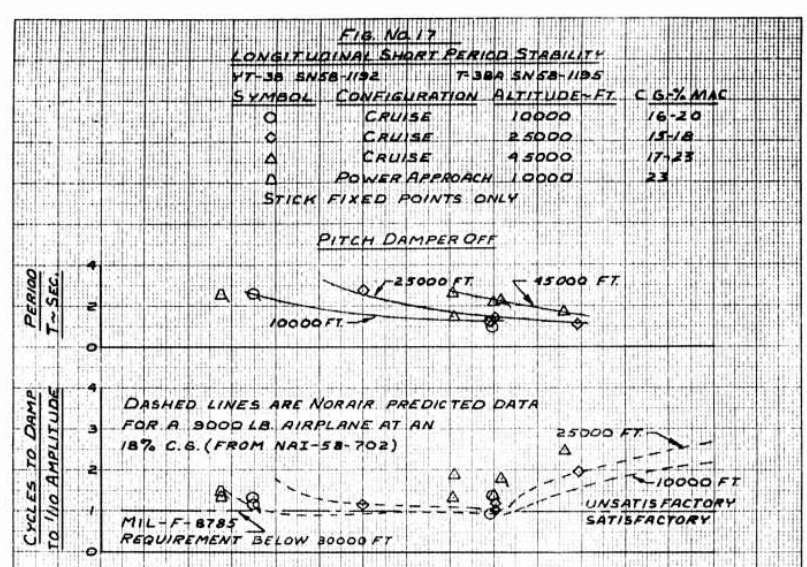}
                \caption{An example of hand-plotted data depicted in the AFFTC test report on the T-38A, from 1961 \cite{dtic1961t38}.}
                \label{fig:AFFTC-TR}
            \end{figure}  

            Figure \ref{fig:SP-table} tabulates the collected data. There were some limitations with these comparison points. Neither the Welborn nor the TPS student data reported dynamic pressure (nor actual air temperature), and so a standard atmosphere was assumed. Also, data were not standardized to a common center of gravity (CG) location; it is known that CG location will alter the short period frequency. Finally the AFFTC report had scanned and hand-plotted data that was difficult to read with precision. An example of a page from the report is shown in figure \ref{fig:AFFTC-TR}. We rejected one point from this source as an outlier.\footnote{The rejected point appeared to be the symbol for 45,000 ft; however it was plotted adjacent to the 25,000 ft and 10,000 ft trend-lines. This outlier is visible in figure \ref{fig:AFFTC-TR} for the unusually interested reader. This point may have had an extreme CG location.}

            \begin{figure}[t]
                \centering
                \includegraphics[width = \linewidth]{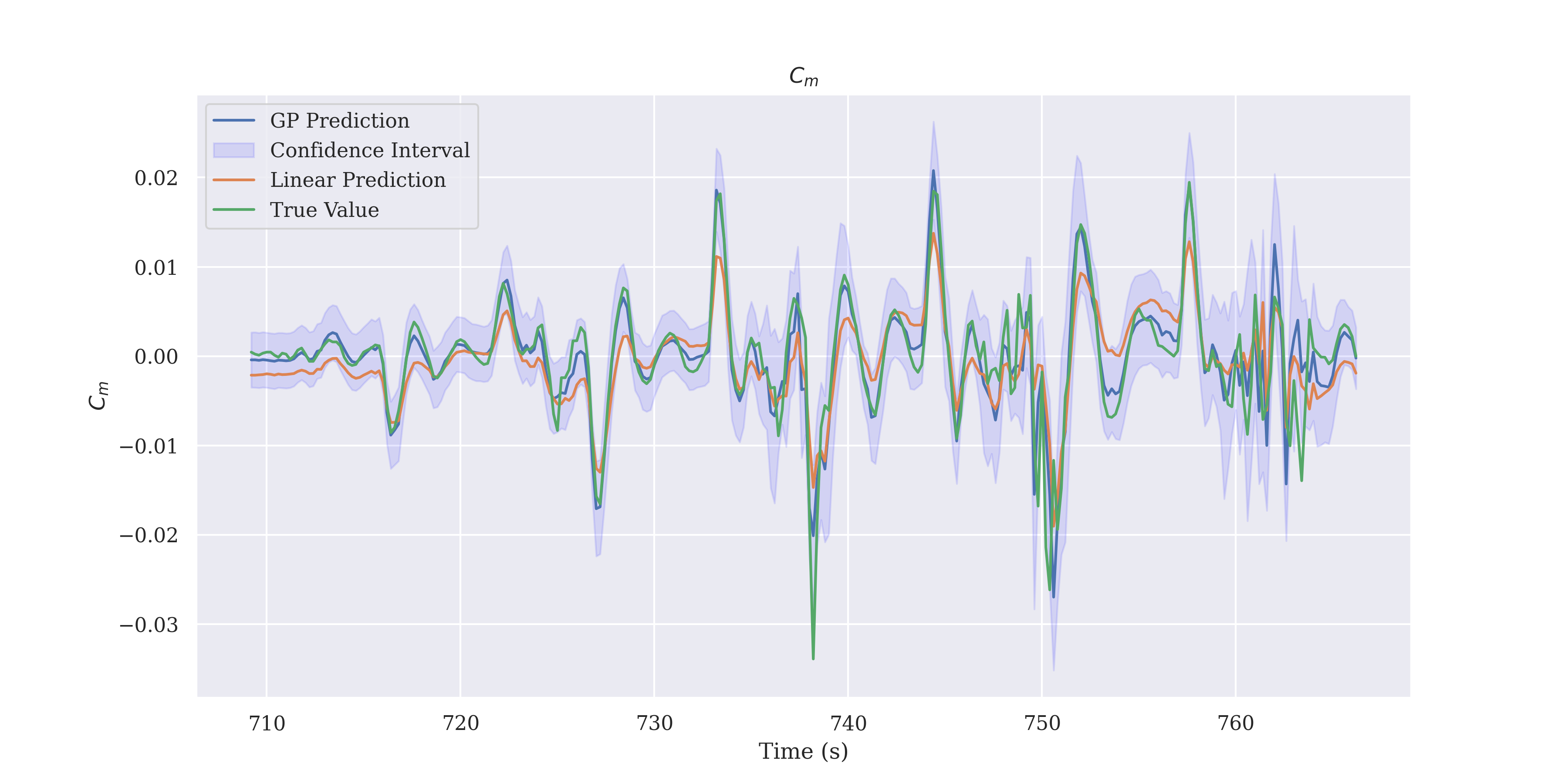}
                \caption{$C_m$ prediction using our Gaussian process based approach. The linear model is constructed using linear regression between the state vector given in equation \ref{eq:state_vector} and $C_m$.}
                \label{fig:cm_pred}
            \end{figure} 

            Figure \ref{fig:cm_pred} illustrates the predictive performance of a Gaussian Process (GP) model for the pitching moment coefficient, \( C_m \), during a rollercoaster maneuver conducted at $0.7$M in the T-38C aircraft. The true values of \( C_m \), calculated using equation \ref{Mbar}, are represented by the dashed line, serving as the ground truth. The solid line denotes the GP prediction, accompanied by a shaded region that captures the 95\% confidence interval, representing the model's uncertainty in its predictions. 
            
            The GP model successfully captures the general trend of \( C_m \), providing reasonably accurate estimates while also quantifying the inherent uncertainty in dynamic maneuvers. This is evidenced by the GP prediction closely following the true moment values throughout the maneuver, with the uncertainty bands effectively encompassing the true values in most regions. Table \ref{tab:cm_comparison} also compares the estimates of the dynamic parameters $C_{m_{\alpha}}, C_{m_{\delta_{e}}}, C_{m_{Q}}, \omega_{SP}, \zeta_{SP}$ between our proposed method, the results from \citeauthor{shepherd2010limited}, and a linear model constructed using linear regression between the state vector given in equation \ref{eq:state_vector} and $C_m$.

            \begin{table}[h!]
            \centering
            \begin{tabular}{|l|c|c|c|c|c|}
                \hline
                Method & $C_{m_{\alpha}}$ & $C_{m_{\delta_{e}}}$ & $C_{m_{Q}}$ & $\omega_{SP}$ & $\zeta_{SP}$ \\
                \hline
                Linear Model   & -0.285 & -0.525  & -3.240  & 0.250 & 0.219 \\
                GP Estimates   & -0.442 & -1.045  & -15.590  & 0.317 & 0.329 \\
                \citeauthor{shepherd2010limited} Estimates \cite{shepherd2010limited} & -0.562   & -1.285    & -12.720    & 0.380     & 0.290     \\
                \hline
            \end{tabular}
            \caption{Comparison of dynamic parameter estimates at $0.7$M and $32,000$ ft. 
            }
            \label{tab:cm_comparison}
        \end{table}

\begin{figure}
                \centering
                \includegraphics[width = 0.9\linewidth]{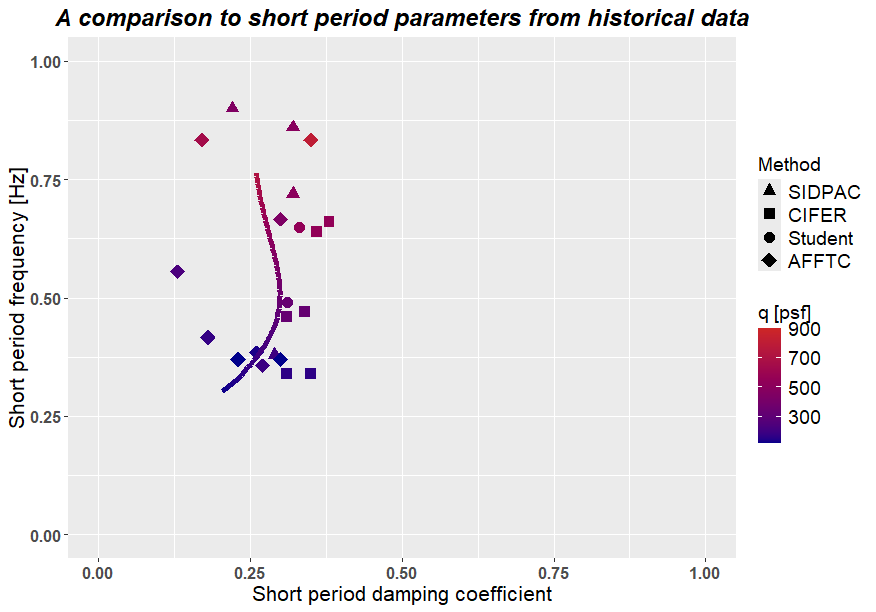}
                \caption{The consolidated short period data collected for comparison, along with a continuous line of values produced using our methods and varying dynamic pressure (denoted as q in this figure).}
                \label{fig:comparison-points}
            \end{figure}

            In figure \ref{fig:comparison-points} we have provided a plot of the short period frequencies and damping coefficients contained in the table of figure \ref{fig:SP-table}. As we have emphasized, the short period frequency is a strong implicit function of dynamic pressure. Since our data-based architecture output a full physics of the short period phenomenon, it allowed us to calculate short period at any dynamic pressure. In figure \ref{fig:comparison-points} we generated a continuous line $(\omega_{SP}, \zeta_{SP}) = \mathcal{F}(\bar{q})$ using our methods and for the full range of dynamic pressures appearing in the comparison data. 

            Qualitatively, the correspondence with the cluster of historical data is extremely good. This is despite the fact that the roller coaster maneuver used to calculate these short period parameters utterly lacked the high frequency content required by SIDPAC, CIFER®, and other SysID methods. In fact, analysis of an exemplar roller coaster revealed that the pilot input AOA oscillations at a rate of around 1.3 deg/sec, or a mere 0.0036 Hz. This frequency input into the system was just 1\% of the average short period frequency in the consolidated comparison data. 
            
            We hypothesize that we were able to reproduce high-frequency modes with low-frequency experimental content due to our imposition of a physics-based prior in the form of the Morelli mean function. While the short period predicted by the A-7 model was completely incorrect, the data ingested by our architecture refined the extant value towards the true one. This exemplifies the importance of using reduced-order modeling methods that trace rigorously to physics-informed HFMs. 


            Our model builder did not explicitly learn the relationship between $\omega_{SP}$ and $\bar{q}$, however it did include $\bar{q}$ as a state variable for the GPR and thus constructed a $\bar{q}$ dependence embedded within the hypersurface. \textit{i.e.,} every time the non-linear forces and moments were calculated, dynamic pressure was taken into account; GPR learned how implicit variations in dynamic pressure affected the forces and moments produced.

            The fact of this embedding distinguishes the data-based architecture from the other methods against which we have compared it. A doublet spot-checks the short period frequency characteristics at one single dynamic pressure. SysID does better, by regressing stability derivatives or outputting transfer functions; nevertheless these also occur at one particular dynamic pressure. Any dependency on $\bar{q}$ is simply not captured. All these other methods require the test pilot to repeat the flight test techniques at a very fine grid of dynamic pressures in order to then regress across that variable. 

\begin{figure}
                \centering
                \includegraphics[width = 0.9\linewidth]{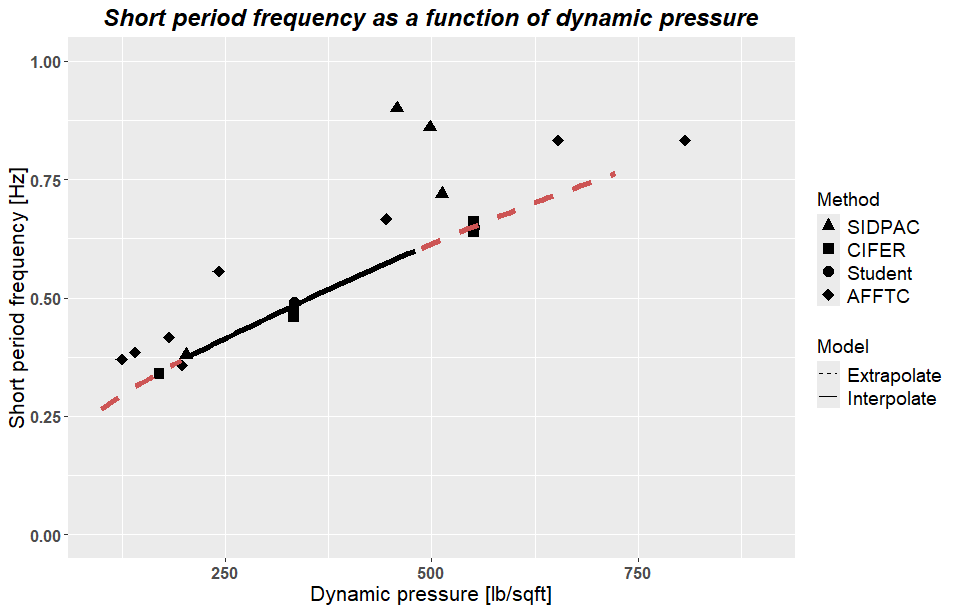}
                \caption{Short period frequency model for dependence on dynamic pressure, along with comparison data. Here, no dependency on Mach was observed.}
                \label{fig:omega-qbar}
            \end{figure}

            Our data-based architecture retrieves this functional dependence with no additional experimental effort. Figure \ref{fig:omega-qbar} plots our result for short period frequency against the comparison data. Since short period frequency weakly depended on Mach, if at all, we picked a moderate Mach value of 0.7 to draw this curve.

            In order to compare our prediction with the internal consistency of the historical record, we subdivided the comparison data into three regions:
            
            \begin{itemize}
                \item A high dynamic pressure region around $515\frac{lb}{ft^2}$.
                \item A moderate dynamic pressure region around $340\frac{lb}{ft^2}$.
                \item A low dynamic pressure region around $165\frac{lb}{ft^2}$.
            \end{itemize}

            Regions of similar dynamic pressure had values for $\bar{q}$ within $\pm40 \frac{lb}{ft^2}$ of each other. This excluded some points from the current analysis, but avoided comparing any historical point to a prediction at a significantly different dynamic pressure. In table \ref{tab:error_metrics}, for each dynamic pressure region we compare the maximum error of our prediction (column "Prediction") against the maximum error internal to the historical record (column "Historical"). The "Difference" column is colored red when the historical record had more internal error than the \emph{worst} error our prediction had at any point; it is colored blue when the opposite is true.

            From table \ref{tab:error_metrics}, our curve predicted the short period frequency to within 5\% of the error internal to the historical record itself. It exceeded the consistency of the historical record for moderate and low dynamic pressure points.

            \begin{figure}
                \centering
                \includegraphics[width = 0.9\linewidth]{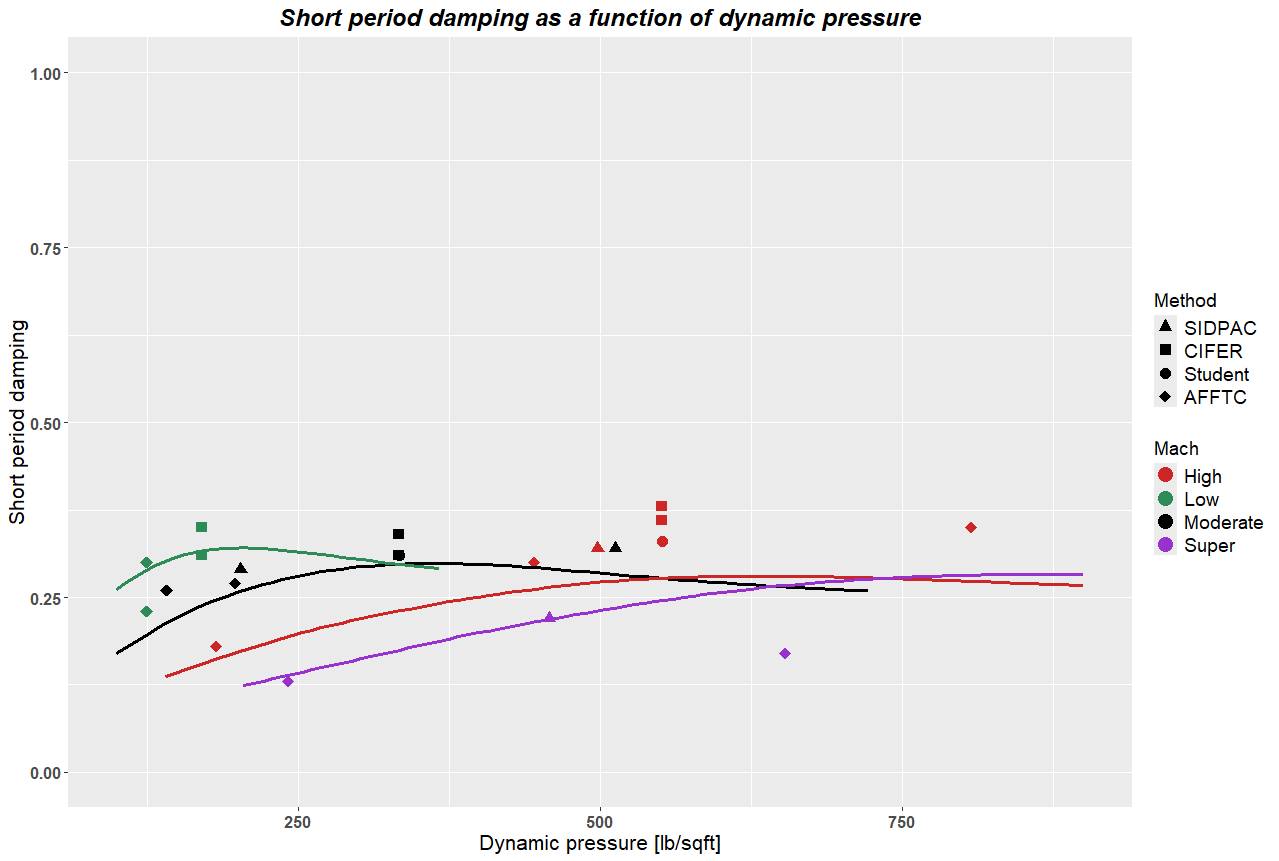}
                \caption{Short period damping model for dependence on dynamic pressure, along with comparison data. Here, different curves demonstrate the Mach dependence.}
                \label{fig:zeta-qbar}
            \end{figure}   
            
            We also computed the short period damping as a function of dynamic pressure, and compare it to the historical record in figure \ref{fig:zeta-qbar}. In the case of damping, we observed variations in short period damping for different Mach numbers. For that reason, we calculated a family of curves for different Mach regions: "low," calculated at 0.5 Mach; "moderate," calculated at 0.7 Mach; "high," calculated at 0.9 Mach; and "supersonic," calculated at 1.08 Mach.  In addition to considering regions of similar dynamic pressure, we also only considered predictions of similar Mach numbers for the analysis of damping. Similar Mach values were within $\pm0.02$.

            
            Table \ref{tab:error_metrics} presents our results for short period damping in each dynamic pressure region, using the same criteria as for the short period frequency analysis, plus the Mach constraint. Our curve predicted the short period damping to within 11\% of the error internal to the historical record itself. It exceeded the consistency of the historical record for low dynamic pressure points. 


            \begin{table}[h!]
            \centering
            \begin{tabular}{lccc|ccc}
            \hline
            \textbf{Dynamic Pressure} & \multicolumn{3}{c|}{$\boldsymbol{\omega_{sp}}$ \textbf{Error (\%)}} & \multicolumn{3}{c}{$\boldsymbol{\zeta_{sp}}$ \textbf{Error (\%)}} \\
             & \textbf{Prediction} & \textbf{Dataset} & \textbf{Difference} & \textbf{Prediction} & \textbf{Dataset} & \textbf{Difference} \\
            \hline
            High      & 30.2 & 25.6 & \myred{-4.6} & 26.3 & 15.8 & \myred{-10.5}\\
            Moderate  & 4.3  & 6.1  & \myblue{+1.8} & 11.8 & 8.8  & \myred{-3.0}\\
            Low       & 14.3 & 19.0 & \myblue{+4.7} & 8.6 & 11.4 & \myblue{+2.8}\\
            \hline
            \end{tabular}
            \caption{Error metrics for $\omega_{sp}$ and $\zeta_{sp}$ across dynamic pressure regions.}
            \label{tab:error_metrics}
            \end{table}

            \section{Conclusions}

            In this paper, we presented a data-based architecture for flight test. This architecture inverts the usual relationship between data and model predictions. Instead of using data to spot-check the predictions of a high-fidelity offline model, our architecture uses data to refine and update a reduced-order model using online methods.

            This architecture has many advantages, but two of them are primary.

            \begin{enumerate}
                \item It eliminates databands and tolerances. Instead of discarding data that does not meet the parameters of the prediction, our method uses that data to update the model at the as-flown conditions. All data are good data. 
                \item It makes the final outcome of a test a compact, physics-based representation of the aircraft under test in the form of a ROM. Unlike a spot-check of an HFM, a self-contained ROM has interpolative, extrapolative, and explanatory power.
            \end{enumerate}
     
            We demonstrated this architecture with a short flight test campaign in the T-38C. Our demonstration used Gaussian Process Regression for the model builder, with a physics-based prior imposed in the form of a mean function created with a high-fidelity wind tunnel database. The output of our model builder was a hypersurface with uncertainty quantified at each point. After specifying a local trim condition, this hypersurface could be differentiated to yield any stability derivative. We used these stability derivatives in a second-order approximation of longitudinal dynamics to predict the short period frequency and damping.

            We also demonstrated the two advantages of the architecture. 
            
            Firstly, no databands or tolerances were applied to the \emph{rollercoaster} maneuvers flown, and we used whatever data we were given. (We did fly near certain altitudes and Mach values in order to adequately compare with the historical record.) 
            
            Second, even though our test campaign did not attempt to model short period dependence on dynamic pressure or Mach, we were able to use the physics embedded within our hypersurface to reproduce both short period frequency and damping as a function of dynamic pressure and Mach. This function predicted the short period frequency to within 5\% of the error internal to the historical record for any dynamic pressure. This function predicted the short period damping to within 11\% of the error internal to the historical record for any dynamic pressure and Mach combination. These values suggest that our single function, generated from five minutes of low-frequency data, is nearly as predictive as the historical record captured in four separate papers spanning 1961 to 2024. 

            The flight test community should continue to investigate, and in some cases adopt, data-based architectures for flight sciences testing.

            \section{Future Work} \label{future}

            The Pointless Project is a multi-disciplinary, multi-agency effort currently funded by the Air Force Office of Scientific Research, AFWERX, and the DoD's Test Resource Management Center. This is its first public result, and much work remains. We want to highlight three upcoming lines of effort.

            First, although GPR with a physics-based prior yielded tantalizingly promising results in practice, we plan a transition to reduced order modeling methods that enforce the underlying physics during the model update itself. We have planned and funded a live-fly experiment on the F-16 as a TPS Test Management Project, which will use the Projection-based reduced order modeling methods introduced by Farhat \emph{et al} \cite{farhat2010prom}. In order to update the model with test data, we will use a non-parametric probabilistic method, which matches measured data and their uncertainties to stochastic representations of the projection-based ROMs \cite{farhat2017npm}. 

            Second, the work of this paper makes it plain that under a data-based architecture, flight test will always require specific types of maneuvers or flight test techniques. For example, our GPR algorithm required dynamical data centered around 1 G, for which we selected a \emph{rollercoaster}. However, with databands and tolerances eliminated, the basic meaning of a maneuver evolves. A maneuver now seeks to generate data of sufficient quality and informational density to reduce the uncertainty on a ROM. We envision maneuvers that optimize \emph{dwells} in some parameter space. For example, the Gaussian Process as presented in this paper might have large uncertainties associated with a particular region of the hypersurface. This wouldn't in general be a coordinate like (Mach, altitude), but instead a multi-dimensional coordinate like (Mach, altitude, pitch rate, AOA). High uncertainty in that region means the test pilot must go gather information at that combination of parameters. One of the activities of test execution will be tracking these regions, and planning FTTs to reduce uncertainty via dwells. 

            Finally, we suggested in section \ref{advantages} that the data-based architecture will enable an outcome-based framework for risk management. However, the methodology for this application has not been fully elaborated. In particular, we are developing statistical classifier methods that allow us to label certain combinations of ROMs in a very high parameter space as "safe" or "unsafe," as required by Jurado \emph{et al} \cite{jurado2022risk}. 

            Significant work is required in each of these areas, and we invite you to join us.

            \section{Acknowledgments}

            The authors would like to thank Stanford's Intelligent Systems Laboratory (SISL) and Mykel Kochenderfer, for expressing interest in this flight test research and allowing Josh to spend many hours laboring on it. 

            And we would like to express copious gratitude to the Test Pilot School for sponsoring our staff TMP and providing us (calibrated!) data. In particular, we extend our thanks to the head of the research division, Chiawei "FUG" Lee, the industrious Jessica "Sting" Peterson, and the TPS Commandant, James "Fangs" Valpiani for his intellectual advocacy for data-based flight test methods. 

            \section{Disclaimer}

            The views expressed in this article, book, or presentation are those of the authors and do not necessarily reflect the official policy or position of the United States Air Force Academy, the Air Force, the Department of Defense, or the U.S. Government.
            
	        \printbibliography

            \newpage

            \appendix
            
            \section{Short Period Transfer Functions} \label{math}

            \subsection{1-DOF Approximation} \label{1-DOF}

            Beginning with the non-linear equations of motion \ref{Xbar} - \ref{Mbar}, we perform a perturbation analysis by assuming small angular deviations from steady, trimmed flight with the center of gravity constrained. Each variable in our analysis becomes the sum of a steady-state value and a perturbative quantity. Lower-case letters represent the perturbation. For example, the aircraft rates become
            
            \begin{align*}
        		\begin{bmatrix}
        			P \\
        			Q \\
        			R \\
        		\end{bmatrix} &=
        		\begin{bmatrix}
        			P_{1}+p \\
        			Q_{1}+q \\
        			R_{1}+r \\
        		\end{bmatrix}
        	\end{align*}
            
            We substitute expressions of this form into equation \ref{Mbar} representing the longitudinal dynamics, and then subtract out the steady-state equation, yielding an ordinary differential equation (ODE) for the perturbation dynamics. In this analysis products of perturbative rates such as $pr$ and $r^2$ are extremely small, so we neglect them. These simplifying assumptions transform equation \ref{Mbar} into the highly manageable ODE:

            \begin{equation} \label{ODE-A}
                \dot{q} = \frac{\bar{M}}{I_{y}} \equiv M
            \end{equation}
            
            We now approximate the sum of the moments as a first-order multivariate Taylor expansion of the perturbation variables that we expect are most important to the dynamics. The approximation used in section \ref{Analytical-Predictions} is
			
			\begin{equation} \label{TaylorSeries-A}
				M = \frac{\partial M}{\partial \alpha}\alpha + \frac{\partial M}{\partial \dot{\alpha}}\dot{\alpha} +
				\frac{\partial M}{\partial \delta_{e}}\delta_{e}
			\end{equation}
          
            In order to investigate the short period, we further assume that all pitch rate  perturbations $q$ are due to small perturbations in the angle of attack $\alpha$. This implies $q \approx \dot{\alpha}$, and we can substitute equation \ref{TaylorSeries-A} into equation \ref{ODE-A}, yielding:
            
            \begin{equation} \label{ODE-AOA-A}
                \ddot{\alpha} = M_{\alpha}\alpha + M_{\dot{\alpha}}\dot{\alpha} + M_{\delta_{e}}\delta_{e} 
            \end{equation}

            Here we have used a subscript to indicate partial differentiation by the sub-scripted variable. 

            Since the short period mode is a response to elevator input, we make the time history of the elevator deflection $\delta_{e}(t)$ a forcing function and take the Laplace Transform of both sides. Rearranging:

            \begin{equation} \label{Laplace-A}
                \frac{\alpha(s)}{\delta_{e}(s)} = \frac{M_{\delta_{e}}}{(s^2 - M_{\dot{\alpha}}s - M_{\alpha})}  
            \end{equation}

            The denominator on the right hand side of equation \ref{Laplace-A} is the characteristic equation for this system. Comparing to the canonical form of the characteristic equation, $s^2 + 2 \zeta \omega s + \omega^2$, we obtain by inspection:\footnote{The familiar reader will object to the omission of the pitch rate damping term, $M_{q}$, in our expression for $\zeta_{SP}$. We have assumed that term is captured within $M_{\dot{\alpha}}$ for simplicity and readability of this section. }

            \begin{align} 
                \omega_{SP} &= \sqrt{-M_{\alpha}} \label{omega-analytical-A} \\
                \zeta_{SP} &= \frac{-M_{\dot{\alpha}}}{2 \sqrt{-M_{\alpha}}} \label{zeta-analytical-A}
            \end{align} 

            \subsection{2-DOF Approximation} \label{2-DOF}
            
            Now we return to the full system \ref{Xbar} - \ref{Mbar} and perform the perturbative analysis above on $\bar{X}$ and $\bar{Z}$. We further assume that products of any rate and the vertical velocity ($w$) or lateral velocity ($w$) are also small. This yields
            
            \begin{equation}
        		\begin{bmatrix}
        			\dot{u} \\
        			\dot{w - U_{1}}q \\
        			\dot{q} \\
        		\end{bmatrix} =
        		\begin{bmatrix}
        			\bar{X}/m \\
        			\bar{Z}/m \\
        			\bar{M}/I_{y} \\
        		\end{bmatrix} \equiv
                \begin{bmatrix}
                    X/m \\
                    Z/m \\
                    M/I_{y} \\
                \end{bmatrix}
        	\end{equation}

            As before, we express each force or moment term as a Taylor series expansion around the variables that we expect to have the strongest impact to the dynamics. Compared to equation \ref{TaylorSeries-A}, we admit a larger number of variables, including forward speed $u$, flight path angle $\theta$, pitch rate $q$, and a perturbative thrust $\delta_{T}$. We continue to assume that angle of attack is small such that $tan(\alpha) \approx \frac{w}{U_{1}} \approx \alpha$. In this case:

            \begin{equation}	
        	\begin{bmatrix}
                \dot{u} \\
                \dot{w} - U_{1}q \\
        	    \dot{q} 
            \end{bmatrix} =
        	\begin{bmatrix}
        	  	-g cos \Theta_{1}\theta + X_{u} u + X_{\alpha}\alpha + X_{\delta_{e}}\delta_{e} + X_{\delta_{T}}\delta_{T} \\
        		-g sin \Theta_{1}\theta + Z_{u} u + Z_{\alpha} \alpha + Z_{\dot{\alpha}} \dot{\alpha} + Z_{q} q + Z_{\delta_{e}}\delta_{e} + Z_{\delta_{T}}\delta_{T} \\
        		M_{u} u + M_{\alpha} \alpha + M_{\dot{\alpha}}\dot{\alpha} + M_{q} q + M_{\delta_{e}}\delta_{e} + M_{\delta_{T}}\delta_{T}
        	\end{bmatrix}
        	\end{equation}

            Here we have explicitly characterized the $X$ and $Z$ force components due to the gravitational vector of magnitude $g$, through the steady state flight path angle $\Theta_{1}$ and its perturbative component $\theta$. 
            
            As before, we pose $\delta_{e}(t)$ as the forcing function and take the Laplace transform of both sides. Rearranging, this yields: 
            
            \begin{equation}
    	 	\begin{bmatrix}
    	 		s - X_{u} - X_{\delta_{T}} & -X_{\alpha} & g cos \Theta_{1} \\
    	 		-Z_{u} & s(U_{1}-Z_{\dot{\alpha}}) - Z_{\alpha} & -(Z_{q}+U_{1})s+ g sin \Theta_{1}) \\
    	 		-(M_{u}+M_{\delta_{T}}) & -(M_{\dot{\alpha}}s + M_{\alpha} + M_{\delta_{T}}) & s^{2}-M_{q}s
    	 	\end{bmatrix}
    	 	\begin{bmatrix}
    	 		\frac{u(s)}{\delta_{e}(s)} \\
    	 		\frac{\alpha(s)}{\delta_{e}(s)} \\
    	 		\frac{\theta(s)}{\delta_{e}(s)} 
    	 	\end{bmatrix} =
    	 	\begin{bmatrix}
    	 		X_{\delta_{e}} \\
    	 		Z_{\delta_{e}} \\
    	 		M_{\delta_{e}} \\
    	 	\end{bmatrix}
            \end{equation}

            Each of the transfer functions of the form $\frac{\alpha(s)}{\delta_{e}(s)}$ follow immediately from this matrix equation via Cramer's rule. To determine the short period in particular, assume that velocity is nearly constant so that $u$ is small compared to $U_{1}$. Assume also with minimal loss of generality that $Z_{\dot{\alpha}} = Z_{q} = \Theta_{1} = M_{\delta_{T}} = 0$. Under these assumptions the transfer function for $\alpha$ becomes

            \begin{align} \label{Computational-TF-A}
        		\frac{\alpha(s)}{\delta_{e}(s)} &= \frac{ \frac{1}{U_{1}} [Z_{\delta_{e}}s+ (M_{\delta_{e}}U_{1} - M_{q}Z_{\delta_{e}})]}{s^{2} - (M_{q} + \frac{Z_{\alpha}}{U_{1}} + M_{\dot{\alpha}})s + (\frac{Z_{\alpha}M_{q}}{U_{1}} - M_{\alpha})}	
        	\end{align}

           As before, we compare the denominator of equation \ref{Computational-TF-A} to the canonical form of a characteristic equation and obtain by inspection
            
            \begin{align} 
		          \omega_{SP} &\approx \sqrt{\frac{Z_{\alpha} M_{q}}{U_{1}} - M_{\alpha}} \label{omega-computational-A} \\
		          \zeta_{SP} &\approx \frac{-1}{2 \omega_{SP}}(M_{q} + \frac{Z_{\alpha}}{U_{1}} + M_{\dot{\alpha}}) \label{zeta-computational-A}
	        \end{align}

            Observe that because the product $Z_\alpha M_{q}$ is small compared to $U_{1}$, equation \ref{omega-computational-A} reduces to the 1-DOF analysis that yielded equation \ref{omega-analytical-A}.
	
	\end{document}